\documentclass[preprint,12pt,3p]{elsarticle}
\usepackage{geometry}

\usepackage{placeins}
\usepackage{xspace}
\usepackage{caption}
\usepackage{subcaption}
\usepackage{graphicx}
\usepackage[hyphens]{url}
\usepackage[hidelinks]{hyperref}
\usepackage{booktabs}
\usepackage{threeparttable}
\usepackage{multirow}
\usepackage{float}
\usepackage{amsmath}

\usepackage{cleveref}
\usepackage{longtable}
\usepackage{natbib}

\usepackage{graphicx}
\usepackage{amssymb}

\usepackage{lineno}
\usepackage[dvipsnames]{xcolor}
\usepackage{ulem}

\definecolor{nyupurple}{RGB}{150, 35, 140}

\journal{arXiv}

\begin{document}

\begin{frontmatter}


\title{Travel Demand Forecasting: A Fair AI Approach}



\author[label1]{Xiaojian Zhang}
\address[label1]{Department of Civil and Coastal Engineering, University of Florida}
\ead{xiaojianzhang@ufl.edu}

\author[label2]{Qian Ke}
\address[label2]{Bloomberg, 731 Lexington Avenue, New York, US}
\ead{qke4@bloomberg.net}

\author[label1]{Xilei Zhao\corref{cor1}}
\ead{xilei.zhao@essie.ufl.edu}

\cortext[cor1]{Corresponding author. Postal address: 1949 Stadium Rd, Gainesville, FL 32611, USA.}

\begin{abstract}
Artificial Intelligence (AI) and machine learning have been increasingly adopted for travel demand forecasting. The AI-based travel demand forecasting models, though generate accurate predictions, may produce prediction biases and raise fairness issues. Using such biased models for decision-making may lead to transportation policies that exacerbate social inequalities. However, limited studies have been focused on addressing the fairness issues of these models. Therefore, in this study, we propose a novel methodology to develop fairness-aware, highly-accurate travel demand forecasting models. Particularly, the proposed methodology can enhance the fairness of AI models for multiple protected attributes (such as race and income) simultaneously. Specifically, we introduce a new fairness regularization term, which is explicitly designed to measure the correlation between prediction accuracy and multiple protected attributes, into the loss function of the travel demand forecasting model. We conduct two case studies to evaluate the performance of the proposed methodology using real-world ridesourcing-trip data in Chicago, IL and Austin, TX, respectively. Results highlight that our proposed methodology can effectively enhance fairness for multiple protected attributes while preserving prediction accuracy. Additionally, we have compared our methodology with three state-of-the-art methods that adopt the regularization term approach, and the results demonstrate that our approach significantly outperforms them in both preserving prediction accuracy and enhancing fairness. This study can provide transportation professionals with a new tool to achieve fair and accurate travel demand forecasting.

\end{abstract}


\end{frontmatter}



\section{Introduction}
\label{S:1}

In recent years, Artificial Intelligence (AI) has been increasingly used in travel demand forecasting, due to its powerful prediction capability \citep{xu2022adaptive,chu2019deep}. However, a growing number of studies reported that AI has evident fairness issues \citep{beutel2019putting, baker2021algorithmic, angwin2016machine, barabas2018interventions, prates2020assessing, buolamwini2018gender, obermeyer2019dissecting}---making worse predictions for disadvantaged population groups (e.g., racial and ethnic minorities, low-income individuals, and women) than the advantaged groups. For example, facial recognition systems have higher error rates on classifying darker-skinned individuals and females \citep{buolamwini2018gender}. Studies in the transportation domain also have similar findings. For example, recent research has shown that AI algorithms could underestimate the actual travel demand for the disadvantaged groups \citep{yan2020fairness} and deliver much lower prediction accuracy for the disadvantaged groups than the advantaged groups \citep{zheng2021equality}. The unfair predictions may negatively impact transportation policies and decision-making (e.g., vehicle rebalancing and traffic control), leading to unintended consequences for transportation equity. Therefore, AI-based travel demand forecasting models should account for both prediction accuracy and fairness \citep{yan2021fairness}.

Recently, some researchers have started to develop fairness-aware AI methods in travel behavior modeling, e.g., travel mode choice modeling \citep{zheng2021equality} and travel demand forecasting \citep{yan2020fairness}. However, research on this important topic, especially for travel demand forecasting, is still lacking. For instance, although various methods have been developed to mitigate the unfairness issues, very few can be flexibly adopted by different types of models (e.g., linear models, deep learning models with different architectures, etc.). In other words, there still lacks a systematic framework to address the model's fairness issue in a model-agnostic (i.e., the method should be independent of models) manner. Also, it remains largely unsolved how to prioritize model fairness while preserving its prediction accuracy, both of which are critical to ensure the trustworthiness of AI \citep{kaur2022trustworthy,li2023trustworthy}. Additionally, previous studies have primarily focused on correcting the unfairness of a single protected attribute. In real-world dataset, however, the debiased model and results could vary across different protected attributes, potentially causing confusion and hindering adoption by end-users. For example, one study has found that mitigating unfairness of one protected attribute (i.e., race) could increase the prediction disparities of another protected attribute (i.e., income) \citep{zheng2021equality}. This suggests that a model that is fair for one protected attribute could still be unfair for other attributes \citep{wan2023processing}. However, few prior studies have been devoted to simultaneously tackling fairness issues from multiple protected attributes \citep{bose2019compositional,wan2023processing}.

To address these research gaps, we aim to develop a new methodology to enhance fairness in AI-based travel demand forecasting models. More specifically, first, we define \textit{Fairness} as the \textit{Equality of Prediction Accuracy}, i.e., the prediction accuracy is equal for advantaged and disadvantaged population groups. Next, we examine the potential unfairness (i.e., prediction accuracy disparity) existing among several state-of-the-art deep learning and statistical models for travel demand forecasting, using real-world ridesourcing-trip data in Chicago, IL and Austin, TX. We propose a novel absolute correlation regularization method to simultaneously correct the detected unfairness across multiple protected attributes (e.g., race, education, etc). We further compare the proposed methodology with other existing state-of-the-art regularization terms to show its effectiveness in both preserving accuracy and correcting unfairness. The unique contributions of this study are presented as follows:

\begin{itemize}
    \item This study is one of the first studies to examine the fairness issues of travel demand forecasting models from the algorithmic view. We extend the literature on this topic by detecting the unfairness issues of several commonly-used deep learning and statistical models and proposing a methodology to correct the unfairness.

    \item  We introduce a novel absolute correlation regularization term to address the model's unfairness arising from \textit{multiple protected attributes}. This regularization term is explicitly designed to penalize models that produce unfair predictions, which holds notable transparency. Moreover, the proposed regularization term is \textit{model-agnostic} and can be flexibly incorporated into the loss function of any type of model architecture.

    \item We propose to use an \textit{interactive} weight coefficient for both the accuracy loss and fairness regularization terms. This weight coefficient is tuned simultaneously with other key hyperparameters of an AI model (e.g., number of hidden layers, number of hidden neurons, and learning rate of a multiple-layer perception model). Therefore, the fairness-aware travel demand forecasting models can optimally improve fairness while preserving prediction accuracy.
\end{itemize}

The remaining paper is structured as follows: Section \ref{LR} reviews the related studies. Section \ref{Methodology} introduces the fairness definitions, metrics and unfairness correction method. We introduce the empirical case studies in Section \ref{data}. The modeling results are presented in Section \ref{results}. Section \ref{discussion} discusses the merits of the proposed methodology, echoes the critical findings, proposes some policy implications and lists several future research directions. Finally, Section \ref{conclusion} concludes our study.

\section{Literature Review}
\label{LR}

\subsection{AI fairness issues}
In recent years, AI methods have been deployed in a broad array of real-world applications due to their outstanding strength in producing highly-accurate predictions. However, there has been a growing recognition that, despite predictive superiority, AI and machine learning techniques have also been accompanied by increasing concerns of fairness \citep{angwin2016machine}. Studies from multiple fields have reported that AI algorithms could be discriminatory to the disadvantaged population groups under various applications, including healthcare, criminal justice, credit assessment, translation, among many others \citep{angwin2016machine, baker2021algorithmic, barabas2018interventions, obermeyer2019dissecting, dressel2018accuracy, prates2020assessing}. For example, healthcare systems could underestimate the health condition of black patients than white patients, even if they have the same health risk score \citep{obermeyer2019dissecting}. If these inherent biases are not addressed, using these AI systems to assist decision-making will worsen the existing social disparities \citep{mehrabi2021survey}.

\subsubsection{Taxonomy of fairness notions}

Numerous fairness notions and corresponding mathematical formulations have been proposed for different downstream learning tasks \citep{mehrabi2021survey}. These fairness notions span various dimensions, including classification vs. regression, group vs. individual and disparate treatment \citep{berk2017convex}. In classification, multiple fairness notions are created to mitigate ``disparate impact'', i.e., if practices or policies have disproportionately adverse effects on different groups \citep{barocas2016big}. For example, \textit{statistical parity} \citep{dwork2012fairness}, \textit{equality of odds} and and \textit{equality of opportunity} \citep{hardt2016equality}. In regression, notions like \textit{individual/region-based fairness gap} \citep{yan2020fairness}, \textit{cross-pair} loss \citep{berk2017convex} and \textit{equal means} \citep{calders2013controlling} are introduced to address real-world regression applications that require fairness concerns. Fairness notions also branch into the axis of individual and group. Individual fairness requires similar individuals to be treated similarly, while group fairness equalizes the outcome among all groups \citep{dwork2012fairness}. Another branch to classify fairness notions is determining whether the \textit{disparate treatment} is allowed. Disparate treatment measures fairness through treatment rather than the outcomes. It addresses both formal classification and intentional discrimination  \citep{barocas2016big}, and includes notions like \textit{counterfactual fairness} \citep{kusner2017counterfactual} and \textit{fairness through unawareness} \citep{dwork2012fairness}. These fairness notions have laid a solid foundation for defining and measuring fairness in real-world problems.

\subsubsection{Correcting unfairness for multiple protected attributes}

There are three possible ways to achieve the aforementioned fairness, i.e., correcting the unfairness. First, \textit{pre-processing} the data (e.g., resampling or reweighting) and remove bias before training the models (e.g., \citep{kamiran2012data, calmon2017optimized}). Second, \textit{in-processing}: modifying the algorithms such as including fairness penalty in the loss function \citep{berk2017convex, yan2020fairness} or incorporating constraints \citep{agarwal2019fair}. Third, \textit{post-processing}: correcting unfairness by adjusting the learned algorithms \citep{johnson2016impartial, hardt2016equality}. In this study, we selected the in-processing techniques due to their transparency (i.e., directly taking fairness into model optimization) and strong capabilities in achieving fairness even when confronted with biased data \citep{caton2020fairness} and the effectiveness in mitigating bias amplification problems (i.e., the trained models amplify the biases in the training data) \citep{wang2021directional}.

In-processing methods involve two categories: implicit method and explicit method \citep{wan2023processing}. Implicit methods debias the models by implicitly removing bias from the latent representations. They usually hypothesize that if the latent representations are less biased, the predictions produced from the representations could also be less biased. The implicit methods are commonly used in adversarial learning \citep{yang2023adversarial, xu2019achieving, yan2021equitensors}, contrastive learning \citep{cheng2021fairfil}, etc. However, these methods (1) are usually less transparent since we can hardly interpret how the produced latent representations mitigate (or even remove) the unfairness \citep{quadrianto2019discovering, du2020fairness} and (2) usually come with specific model architectures \citep{yan2021equitensors}. Explicit methods focus on explicitly modifying the objective function while keeping the model structure intact, for example, adding fairness-related regularization terms or constraints. Therefore, the explicit methods usually afford greater flexibility and can be applied to a wide range of models. Existing explicit methods include absolute correlation regularization term \citep{beutel2019putting}, pairwise fairness loss \citep{berk2017convex}, equal means \citep{calders2013controlling}, etc. This study adopts the explicit method by integrating a fairness-related regularization term into the loss function to jointly account for accuracy and fairness.

Achieving multi-attribute fairness has long been an enduring challenge in using in-processing techniques to mitigate unfairness \citep{wan2023processing}. To date, most of the existing literature purely focused on correcting the unfairness of a single protected attribute \citep{berk2017convex, yang2023adversarial,kamishima2011fairness,agarwal2019fair}. However, mitigating the unfairness of one attribute may increase the unfairness of another attribute \citep{zheng2021equality}. This unexpected outcome may confuse the end-users (e.g., travel demand modelers) and thus hinder the adoption of the fairness-aware models. To tackle this issue, \citet{yan2020fairness} proposed to explicitly correct the unfairness of multiple attributes by simply adding multiple regularization terms (one for each attribute with a corresponding weight) into the loss function. However, when the protected attributes are correlated with each other (which is the case for most travel demand forecasting problems), it could be challenging to determine the appropriate weight for each protected attribute in order to achieve the optimal solution that minimizes the unfairness for the combination of the selected protected attributes. Other related methods include learning fair graph embeddings via adversarial learning \citep{bose2019compositional}, disentangled representation learning \citep{kim2021counterfactual}, adding fairness constraints for each protected attribute and achieve fairness via constrained optimization \citep{kearns2018preventing, kearns2019empirical}. However, as we discussed, these methods are often less transparent and come with specific model architectures, which hinder their adaptability. As of now, there is a pressing need to develop transparent, effective and flexible methods that can simultaneously account for fairness for multiple protected attributes and can be applied to any model class.



\subsection{Addressing AI fairness issues in travel demand forecasting}
Recently, transportation researchers have also started to examine and address the fairness concerns of travel demand forecasting models, e.g., \citet{yan2020fairness} and \citet{yan2021equitensors}. Specifically, \citet{yan2020fairness} treated fairness as equal mean per capita travel demand across groups over a period of time and evaluated the fairness issues of several AI methods on demand prediction for ridesourcing services and bike-share systems. Results showed that machine learning spontaneously underestimated the travel demand of disadvantaged people. They also proposed two fairness regularization terms and a corresponding fairness-aware demand prediction model to correct the unfairness. \citet{yan2021equitensors} proposed to use an implicit method, which contains fair representations (i.e., EquiTensors) learned by adversarial learning, to forecast the bike-share demand. These fairness-aware models offer transportation professionals new insights on transportation resource allocations and a novel instrument for designing a fairer transportation ecosystem. 

However, there are still two critical knowledge gaps that have yet to be addressed. Firstly, prior research has primarily concentrated on equalizing per capita travel demand among different population groups, but we should note that travel demand disparities may have already been introduced during the data creation process, which is often beyond our control \citep{zheng2021equality, chouldechova2018frontiers}. For example, multiple studies found that rich people are more likely to use ridesourcing services than the poor \citep{yan2020using,zhang2022machine}. That means this behavioral bias among different population groups may naturally exist \citep{olteanu2019social}. However, to date, no study has investigated how to appropriately account for this type of bias, especially for travel demand forecasting models. Second, the existing fairness-aware travel demand forecasting methods necessitate particular model structures, which has very limited adaptability. Thus, developing a model-agnostic (i.e., independent of the model structure) method that can be flexibly adopted by different types of AI models is promising. To date, however, a systematic method in model-agnostic manner to address fairness issues, especially for travel demand forecasting problems, is still lacking.


\section{Methodology}
\label{Methodology}

The methodological framework is outlined as follows. The travel demand forecasting problem will be mathematically defined in Section \ref{3.1}. In Section \ref{3.2}, we will introduce the fairness metrics used in the proposed methodology, followed by the unfairness correction approach for multiple attributes (in Section \ref{3.3}). The notations are summarized in Table \ref{tab:notations}.

\begin{table}[H]
\centering
\caption{A list of symbols and notations}
\label{tab:notations}
\resizebox{\textwidth}{!}{%
\begin{tabular}{@{}llll@{}}
\toprule
Notations                 & Description                                                                                                  & Notations                    & Description                                                                                                                                                                                  \\ \midrule
\textit{Indices and Sets} &                                                                                                              & $L$                 & overall loss function                                                                                                                                                                        \\
$G$                       & graph                                                                                                        & $l$                 & primary loss function for forecasting model                                                                                                                                                  \\
$V$                       & the set of nodes                                                                                             & \textit{Variables}           &                                                                                                                                                                                              \\
$T$                       & the set of time                                                                                              & $d_{i,j}$           & the distance between node $i$ and node $j$                                                                                                                                                   \\
$\mathcal{J}$             & the index set of attributes                                                                                  & $z_j^i$             & the value of the protected attribute $j$ at node $i$                                                                                                                                 \\
$\mathcal{I}$             & the index set of nodes                                                                                       & $\boldsymbol{z}_j$               & the protected attribute $j$, $\boldsymbol{z}_j = (z_j^i, i \in \mathcal{I})$                                                                                                                                                           \\
$\mathcal{I}_j^{+}$       & the set of advantaged node index                                                                               & $x_t^i$                      & travel demand at node $i$ at time $t$                                                                                                                                    \\
$\mathcal{I}_j^{-}$       & the set of disadvantaged node index                                                                               & $\widehat{x}_t^i$                & estimated travel demand of node $i$ for time $t$                                                                                                                                             \\
$t$                       & timestamp                                                                                                    & $\boldsymbol{x}_t$           & ground truth travel demand at time $t$, $\boldsymbol{x}_t = (x_t^i, i \in \mathcal{I})$                                                                     \\
$E$                       & \begin{tabular}[c]{@{}l@{}}the set of edges representing the connectivity \\  between two nodes\end{tabular} & $\widehat{\boldsymbol{x}}_t$ & estimated travel demand at time $t$, $\widehat{\boldsymbol{x}}_t = (\widehat{x}_t^i, i \in \mathcal{I})$                                                                                                                         \\
$w_{i,j}$                 & the element in the weighted adjacency matrix                                                                 & $\boldsymbol{Y}_t$           & \begin{tabular}[c]{@{}l@{}}ground truth travel demand of next $M$ time intervals \\ starting from $t$, $\boldsymbol{\widehat{Y}}_t = [\boldsymbol{\widehat{x}}_{t}, \cdots, \boldsymbol{\widehat{x}}_{t+M-1}]$ \end{tabular}                                                                                                                      \\
$\boldsymbol{W}$                       & weighted adjacency matrix                                                                                    & $\widehat{\boldsymbol{Y}}_t$              & \begin{tabular}[c]{@{}l@{}}ground truth travel demand of next $M$ time intervals \\ starting from $t$, $\boldsymbol{\widehat{Y}}_t = [\boldsymbol{\widehat{x}}_{t}, \cdots, \boldsymbol{\widehat{x}}_{t+M-1}]$ \end{tabular}                                                                                                                                                                                          \\
$N_j^+$                   & the size of the set of advantaged node index                                                                   & $\boldsymbol{X}_{t}$         & \begin{tabular}[c]{@{}l@{}} input historical $K$ travel demand\\ before time $t$, $\boldsymbol{X}_t = \left[ \boldsymbol{x}_{t-K}, \cdots, \boldsymbol{x}_{t-1} \right]$ \end{tabular}                                                                                                                                                                                       \\
$N_j^-$                   & the size of the set of disadvantaged node index                                                                   & $\boldsymbol{Z}$             & the matrix of protected attributes, $\boldsymbol{Z} = [\boldsymbol{z}_j, j \in \mathcal{J}]$                                                                                                                                                          \\
\textit{Parameters}       &                                                                                                              & $p_j^i$                      & \begin{tabular}[c]{@{}l@{}}the binary indicator indicating if node $i$ is belonging to advantaged\\ ($p_j^i=1$) or disadvantaged ($p_j^i=0$) groups for protected attribute $j$\end{tabular} \\
\textit{$K$}              & length of input historical sequence                                                                          & $\boldsymbol{e}_t$           & the prediction accuracy at time $t$, $\boldsymbol{e}_t = (e_t^i, i \in \mathcal{I})$                                                                                                                                                                                       \\
$M$                       & length of output sequence                                                                                    & $e_t^i$                      & the prediction accuracy of node $i$ at time $t$                                                                                                                                              \\
$N$                       & the number of nodes                                                                                          & $r(\boldsymbol{e}_t, \boldsymbol{z}_j)$                & \begin{tabular}[c]{@{}l@{}}the correlation between prediction accuracy at time $t$ and\\ the protected attribute $z_j$\end{tabular}                                                           \\
$\lambda$                 & interactive weight coefficient                                                                               & $R(\boldsymbol{e}_t,\mathbf{Z})$          & \begin{tabular}[c]{@{}l@{}}multiple correlation coefficient between prediction accuracy at time $t$\\ and a set of protected attributes\end{tabular}                                                     \\
$Q$                       & the total number of protected attributes                                                                     & $\bar{\boldsymbol{e}}_t$                  & the expectation of prediction accuracy $\boldsymbol{e}_t$                                                                                                                                               \\
\textit{Functions}        &                                                                                                              & $\bar{\boldsymbol{z}}_j$                  & the expectation of protected attribute $\boldsymbol{z}_j$                                                                                                                                                 \\
$h(\cdot)$                & function of the travel demand forecasting problem                                                            &                              &                                                                                                                                                                                              \\ \bottomrule
\end{tabular}%
}
\end{table}


\subsection{Travel demand forecasting problem}
\label{3.1}

The goal of travel demand forecasting is to predict the future travel demand for each area (or other spatial unit such as traffic segments) given previously observed time-series data. We consider the transportation network as a weighted directed graph $G=(V, E, \boldsymbol{W})$, where $V$ is a set of nodes (i.e., areas or traffic segments) with $|V| = N$; $E$ is a set of edges representing the connectivity between two nodes; and $\boldsymbol{W} \in \mathbb{R}^{N \times N}$ is a weighted adjacency matrix representing the node's proximity (e.g., distance or functional similarity). Given weighted directed graph $G$ with $N$ nodes, we assume time $t \in T$ is a discrete variable where $T$ is a set containing all possible timestamps, let $\boldsymbol{x}_t = (x_t^i, i \in \mathcal{I})$ represent travel demand at time $t$, where $\mathcal{I}$ is the index set of nodes, $x_t^i$ is the travel demand corresponding to node $i \in \mathcal{I}$ at time $t$, and let $\boldsymbol{X}_t = \left[ \boldsymbol{x}_{t-K}, \cdots, \boldsymbol{x}_{t-1} \right]$ be historical $K$ travel demand before $\boldsymbol{x}_t$. The travel demand forecasting problem could be formulated as learning a function $h(\cdot): \mathbb{R}^{N \times K} \to \mathbb{R}^{N \times M}$ which maps the historical $K$ travel demand to travel demand at next $M$ time interval for all nodes in a given graph $G$. Let $\boldsymbol{\widehat{Y}}_t = [\boldsymbol{\widehat{x}}_{t}, \cdots, \boldsymbol{\widehat{x}}_{t+M-1}]$  denote the predicted travel demand for next $M$ time interval starting from timestamp $t$, where $\boldsymbol{\widehat{x}}_{t} = (\widehat{x}_t^i, i \in \mathcal{I})$ refers to the predicted travel demand at timestamp $t$ for all nodes, then we can mathematically write: 

\begin{equation}
h\left(\boldsymbol{X}_t\right | G) =  \boldsymbol{\widehat{Y}}_t = [\boldsymbol{\widehat{x}}_{t}, \cdots, \boldsymbol{\widehat{x}}_{t+M-1}]
\end{equation}

\subsection{Fairness in travel demand forecasting models} 
\label{3.2}

This study defines \textit{\textbf{Fairness}} as the \textit{\textbf{equality of prediction accuracy}}. Intuitively, we assume that the travel demand prediction accuracy should be independent of the protected attributes. Taking racial composition as an example, equality of prediction accuracy suggests that the prediction accuracy for any racial group should be equal. In this study, we use the Absolute Percentage Error (APE) to measure the predictive accuracy instead of the Mean Absolute Error (MAE) or Root Mean Square Error (RMSE). We believe the magnitude of the travel demand (especially for the emerging mobility) for an advantaged community (e.g., high-income community) should be naturally greater than a disadvantaged community \citep{brown2019prevalence}. This type of behavioral bias may largely be introduced during the data creation process instead of applying the algorithm \citep{olteanu2019social, chouldechova2018frontiers}. If we quantify the equality of prediction accuracy with MAE and RMSE, which are sensitive to the magnitude of the forecasting outcome, machine learning may replicate, or even reinforce and potentially exacerbate existing biases. Instead, APE scales the magnitude and cancels out the behavioral bias that has already been embedded in the data.

Recall from the previous section, a travel demand forecasting model is to learn a function $h$ which takes $K$ historical travel demands $\left[ \boldsymbol{x}_{t-K}, \cdots, \boldsymbol{x}_{t-1} \right]$ as input and predict travel demands from next $M$ time interval starting from time $t$, i.e., $\boldsymbol{Y}_{t}$. We define $\boldsymbol{e}_t = (e_t^i, i \in \mathcal{I})$ to indicate the prediction accuracy (i.e., APE) at time $t$, and $e_t^i$ is the prediction accuracy of node $i$ at time $t$. Specifically,  

\begin{linenomath}
\begin{align*}
 e_t^i&= \left|\frac{x^i_t-\hat{x}^i_t}{x^i_t}\right|
\end{align*}
\end{linenomath}

\noindent where $x^i_t$, $\widehat{x}^i_t$ are the ground truth and predicted value of node $i$ at time $t$, respectively, $e_t^i$ is the absolute percentage error for node $i$ at time $t$. The lower the value of $e_t^i$, the better the predictive performance. 


Suppose $\boldsymbol{Z} = [\boldsymbol{z}_j, j \in \mathcal{J}]$ is the matrix of protected attributes of interest, where $\mathcal{J} = \{1,2, \ldots, Q\}$ is the index set of attributes, where $Q$ is the total number of protected attributes; $\boldsymbol{z}_j = (z_j^i, i \in \mathcal{I})$ represents the protected attribute $j$, and $z_j^i$ denotes the protected attribute $j$ at node $i$, $\mathcal{I}$ is the set of index for nodes. Denote $p_j^i$ as a binary indicator indicating if node $i$ is belonging to advantaged (i.e., $p_j^i = 1$) or disadvantaged (i.e., $p_j^i = 0$) groups for protected attribute $j$, and accordingly let $\mathcal{I}_j^{+} = \{i:  p_j^i = 1\}$ and $\mathcal{I}_j^{-} = \{i: p_j^i = 0\}$ represent the set of advantaged and disadvantaged node index for demographic attribute $j$ with size $N_j^+ =|\mathcal{I}_j^{+}|$ and $N_j^- =|\mathcal{I}_j^{-}|$, respectively. We note that assigning value for $p_j^i$, i.e., determining whether each node should be labeled as advantaged or disadvantaged, is context-specific. This determination could be guided by the criteria or statistics defined by local government \citep{yan2020fairness}. Subsequently, \textit{\textbf{Equality of Prediction Accuracy}} is defined as:

\begin{linenomath}
\begin{align*}
\label{equality of prediction accuracy}
\mathbb{E}\left(\boldsymbol{e}_t \vert p_j^i = 1 \right) &= \mathbb{E}\left(\boldsymbol{e}_t \vert p_j^i = 0 \right)  \quad \forall j \in \mathcal{J}, \forall i \in \mathcal{I}  \\
i.e. \quad \frac{1}{N_j^+ } \cdot \sum_{i \in \mathcal{I}_j^+} e_t^i & =  \frac{1}{N_j^-} \cdot \sum_{i \in \mathcal{I}_j^-} {e_t^i} \quad \forall  j\in \mathcal{J}
\end{align*}
\end{linenomath}

\noindent Where $\mathbb{E}\left(\boldsymbol{e}_t \vert p_j^i = 1 \right)$  and $\mathbb{E}\left(\boldsymbol{e}_t \vert p_j^i = 0 \right)$ are the conditional expectation of prediction accuracy $\boldsymbol{e}_t$ given $p_j^i = 1$ and $p_j^i = 0$, and represents the mean APE for advantaged group and disadvantaged group respectively. That is, for any protected attribute $j$, a fair model should have equal prediction accuracy for different groups. Moreover, when a forecasting model is conducted, we could measure the model fairness by quantifying prediction accuracy disparities, especially between nodes with different labels, for instance, low-income communities and high-income communities. 

In this study, we introduce \textit{\textbf{Prediction Accuracy Gap}} (\textbf{PAG}) as a fairness metric to measure prediction accuracy disparity and if fairness/unfairness achieves/occurs. Define: 

\begin{linenomath}
\begin{equation}
    {PAG}_j = \mathbb{E}\left(\boldsymbol{e}_t \vert p_j^i = 0 \right) - \mathbb{E}\left(\boldsymbol{e}_t \vert p_j^i = 1 \right), \forall i \in \mathcal{I}
\end{equation}
\end{linenomath}

Intuitively speaking, \textit{PAG} directly measures the prediction accuracy disparity between these two types of nodes. A high value of \textit{PAG} indicates that the machine learning model delivers inconsistent predictive performance among nodes; in most cases, the performance is worse in disadvantaged nodes.


In this study, we also use \textit{\textbf{Correlation Coefficient}} as another fairness metric. The correlation coefficient can naturally measure the extent to which the predictions are biased on specific protected groups. Intuitively, if fairness is achieved, correlation between prediction accuracy and any protected attribute should be zero. By using correlation coefficient as a measure of fairness, we assume that the target variable (i.e., prediction accuracy) is linearly correlated with the independent variable (i.e., protected attribute).

Recall from the discussions above, $\boldsymbol{e}_t$ is the prediction accuracy (APE) at time $t$, and $\boldsymbol{z}_t$ refers to the protected attribute $j$ for all nodes. Then, the correlation between prediction accuracy $\boldsymbol{e}_t$ and the protected attribute $\boldsymbol{z}_j$ across all nodes is denoted by $r(\boldsymbol{e}_t, \boldsymbol{z}_j)$. Define: 

\begin{linenomath}
\begin{equation}
\label{corr}
r(\boldsymbol{e}_t, \boldsymbol{z}_j)=\frac{\sum_{i \in \mathcal{I}}\left(e^i_t-\bar{\boldsymbol{e}}_t\right)\left(z^i_j-\bar{\boldsymbol{z}}_j\right)}{\left(\sqrt{\sum_{i \in \mathcal{I}}\left(e^i_t-\bar{\boldsymbol{e}}_t\right)^{2}}\right) \left(\sqrt{\sum_{i \in \mathcal{I}}\left(z^i_j-\bar{\boldsymbol{z}}_j\right)^{2}}\right)}
\end{equation}
\end{linenomath}

\noindent where $\bar{\boldsymbol{e}}_t = \mathbb{E}(\boldsymbol{e}_t)$ and $\bar{\boldsymbol{z}}_j = \mathbb{E}(\boldsymbol{z}_j)$. In our experiment, we add small $\epsilon = e^{-20}$ to denominator to keep it always positive. Although correlation coefficient does not require a label for each region, we cannot directly read the prediction accuracy disparity from it. 

\subsection{Unfairness correction method for travel demand forecasting models}
\label{3.3}
In this study, we introduce an absolute correlation regularization approach, which adapts the efforts from \citet{beutel2019putting}, to mitigate the prediction accuracy disparities existing among groups. In \citet{beutel2019putting}, the authors applied this approach to a classification problem by minimizing the false positive rate (FPR) gap between groups. We generalize this approach to a regression setting (i.e., travel demand forecasting problem) by minimizing the prediction accuracy disparities among different communities.

More importantly, including \citet{beutel2019putting}, most previous studies have primarily focused on correcting the unfairness of one single attribute. In real-world dataset, however, the debiased model and results could differ among various protected attributes. Also, a model that is fair for one protected attribute could still be unfair for other attributes \citep{wan2023processing, zheng2021equality}. One feasible solution to solve this issue is to consider multiple attributes at the same time when correcting the unfairness of the models. We expected that a fair model should produce fair predictions for all types of attributes instead of focusing solely on one. 

Therefore, we propose a methodology that can correct the unfairness for multiple protected attributes. More specifically, we propose to use the \textbf{\textit{Multiple Correlation Coefficient}} \citep{BAI200355}, denoted as $R$, to measure the correlation between the target variable, i.e., prediction accuracy, and a set of protected attributes (including race, education, age and income). A larger $R$ suggests that a stronger dependence may exist between the target variable and the explanatory variables. We expect that a fair prediction should lead to $R = 0$, or at least, a small value. Accordingly, we will use $R$ as the regularization term in the loss function to account for fairness loss. We should note that the linear model may encounter potential multicollinearity concerns. However, there is no need to address them since the goal of the linear model is forecasting rather than estimating the coefficients \citep{shmueli2010explain}.

Recall from previous subsections, we will use the prediction accuracy $\boldsymbol{e}_t$ as the target variable and $\boldsymbol{Z} = [\boldsymbol{z}_j, j \in \mathcal{J}]$ to represent the matrix of multiple protected attributes of interest. And, we use $r(\boldsymbol{e}_t, \boldsymbol{z}_j)$ to indicate the correlation between prediction accuracy $\boldsymbol{e}_t$ and the protected attribute $\boldsymbol{z}_j$ across all nodes. Given these notations, we will naturally write the vector of correlations between each protected attribute $\boldsymbol{z}_j$ and prediction accuracy $\boldsymbol{e}_j$, i.e., $\boldsymbol{c}=\left(r(\boldsymbol{e}_t, \boldsymbol{z}_1), r(\boldsymbol{e}_t, \boldsymbol{z}_2), \ldots, r(\boldsymbol{e}_t, \boldsymbol{z}_Q)\right)^{\top}$, and the correlation matrix calculated by the correlation coefficient among each pair of protected attributes, denoted as $\mathbf{\Omega}$, i.e.,

$$
\mathbf{\Omega}=\left(\begin{array}{cccc}
r(\boldsymbol{z}_1, \boldsymbol{z}_1) & r(\boldsymbol{z}_1, \boldsymbol{z}_2) & \ldots & r(\boldsymbol{z}_1, \boldsymbol{z}_Q) \\
r(\boldsymbol{z}_2, \boldsymbol{z}_1) & \ddots & & \vdots \\
\vdots & & \ddots & \\
r(\boldsymbol{z}_Q, \boldsymbol{z}_1) & \ldots & & r(\boldsymbol{z}_Q, \boldsymbol{z}_Q)
\end{array}\right)
$$

Consequently, the multiple correlation coefficient between $\boldsymbol{e}_t$ and $\mathbf{Z}$), i.e., $R(\boldsymbol{e}_t,\mathbf{Z})$, which is the square root of the coefficient of determination (i.e., $R^2$) of the linear model \citep{allison1999multiple}, can be written as:

\begin{equation}
\label{R}
R(\boldsymbol{e}_t,\mathbf{Z})= \sqrt{\boldsymbol{c}^{\top} \mathbf{\Omega^{-1}} \boldsymbol{c}},
\end{equation}

\noindent where $\boldsymbol{c}^{\top}$ is the transpose of $\boldsymbol{c}$ and $\mathbf{\Omega^{-1}}$ is the inverse matrix of $\mathbf{\Omega}$.

Accordingly, given graph $G$ and a forecasting model $\boldsymbol{\widehat{Y}}_t = h (\boldsymbol{X}_t|G)$, we add the multiple correlation coefficient, $R$, into the loss function, denoted as $L(\boldsymbol{X}_t, \mathbf{Z}|G )$ as shown in Eq. \ref{multi_attribute_loss}. In this way, the model will simultaneously account for the unfairness issues sourcing from multiple protected attributes. Let $\boldsymbol{Y}_t = [\boldsymbol{x}_{t}, \cdots, \boldsymbol{x}_{t+M-1}]$ denote the ground truth travel demand of next $M$ time intervals starting from $t$, mathematically, the loss function of the forecasting model to be minimized, i.e., $L(\boldsymbol{X}_t, \mathbf{Z}|G )$, is written as:


\begin{linenomath}
\begin{equation}
\label{multi_attribute_loss}
L(\boldsymbol{X}_t, \mathbf{Z}|G ) =  \sum_t \left\{\left(1-\lambda\right)\cdot l\left(\boldsymbol{Y}_{t}, h (\boldsymbol{X}_t|G)\right)+\lambda R(e_t,\mathbf{Z}) \right\},
\end{equation}
\end{linenomath}

\noindent and,

\begin{linenomath}
\begin{equation}
\label{mse}
l\left(\boldsymbol{Y}_{t}, h (\boldsymbol{X}_t|G)\right)=\frac{1}{N} \sum_{i=1}^{N} \left(x^i_t - \widehat{x}^i_t\right)^{2}
\end{equation}
\end{linenomath}

\noindent In the above equations, $x^i_t$, $\widehat{x}^i_t$ refer to the ground truth and predicted travel demand for node $i$ at time $t$, respectively; $l$ is the primary loss function for forecasting model, and in this study, we use mean squared error (MSE) for $l$; $\lambda$ is the \textit{interactive} weight coefficient that controls the weight between the prediction loss and the fairness regularization term. When $\lambda = 0$, the model will be unaware of the fairness; and when $\lambda = 1$, the model will completely focus on correcting the unfairness. We can directly treat $\lambda$ as a hyperparameter to find the optimal model that effectively addresses fairness while preserving accuracy. The prediction accuracy disparity is captured and mitigated by the correlation regularization term, in Eq. (\ref{R}). The regularization term is dedicated to shrinking the potential prediction accuracy disparity that existed among groups toward zero. Incorporating it into the loss function enables the machine learning model to automatically keep track of the fairness during training. 


Note that when there is only one single protected attribute of interest, the multiple correlation coefficient, i.e., Eq. \ref{R} reduces to Eq. \ref{corr}.

\section{Case Study}
\label{data}

In this section, we will describe two real-world ridesourcing-trip datasets and seven commonly-used travel demand forecasting models used for case studies. Section \ref{chicago_data} and Section \ref{austin_data} present the data collection and processing process. Table \ref{tab:descriptive statistics} presents the descriptive statistics of all input variables. In Appendix.A, Fig. \ref{fig:ridesourcing demand} displays the spatial distribution of the average ridesourcing demand per hour. We will briefly introduce the selected deep learning and statistical models for unfairness detection and correction in Section \ref{benchmark models}.

\begin{table}[!ht]
\centering
\caption{Descriptive Statistics}
\label{tab:descriptive statistics}
\resizebox{\textwidth}{!}{%
\begin{tabular}{@{}lllllllll@{}}
\toprule
                                                              & \multicolumn{4}{l}{Chicago}       & \multicolumn{4}{l}{Austin}      \\ \cmidrule(l){2-9} 
                                                              & Min  & Max     & Mean  & St. Dev. & Min  & Max    & Mean & St. Dev. \\ \midrule
\textit{Target Variable}                                      &      &         &       &          &      &        &      &          \\
Ridesourcing hourly demand                                    & 0.00 & 2150.00 & 12.01 & 41.26    & 0.00 & 820.00 & 1.41 & 8.81     \\
\textit{Demographic Characteristics}                          &      &         &       &          &      &        &      &          \\
\textbf{Race}: Percentage of white population                          & 0.00 & 0.97    & 0.46  & 0.33     & 0.44 & 0.99   & 0.76 & 0.13     \\
\textbf{Edu}: Percentage of population with a bachelor's degree or above & 0.01 & 0.95    & 0.35  & 0.26     & 0.06 & 0.92   & 0.49 & 0.22     \\
\textbf{Age}: Percentage of young population (aged 18 - 44)            & 0.21 & 0.90    & 0.44  & 0.12     & 0.14 & 0.98   & 0.48 & 0.14     \\
\textbf{Income}: Percentage of low-income households                   & 0.02 & 0.84    & 0.29  & 0.16     & 0.01 & 0.84   & 0.18 & 0.12     \\ \bottomrule
\end{tabular}%
}
\end{table}

\subsection{Chicago ridesourcing-trip data}
\label{chicago_data}

In this study, we collected the publicly available ridesourcing-trip data from Chicago Data Portal\footnote{\url{https://data.cityofchicago.org/Transportation/Transportation-Network-Providers-Trips-2018-2022-/m6dm-c72p/explore}} for case study. The data are from November 1, 2018 to March 31, 2019, containing 45,338,599 trips. There are plenty of attributes included in this dataset, but only pick-up locations and timestamps are considered for this research. Since we focused on trip generation (i.e., origin demand) forecasting, all trips are aggregated at the census-tract level and hourly counted. We prepared the data for modeling in the same way as previous studies \citep{zhang2022machine}, to account for the missing-data issues and outliers. The data preparation process produced the trip generation data for 711 census tracts. We split the first 70\% data for training, the following 10\% for validation and the remaining for testing. The census-tract-level demographic data (i.e., protected attributes) were collected from the American Community Survey 2013–2017 5-year estimates data, including the percentage of white, the percentage of low-income households, the percentage of population with a bachelor's degree or above and the percentage of young populations (with age in 18-44).

\subsection{Austin ridesourcing-trip dataset}
\label{austin_data}
This study also collected ridesourcing-trip data from RideAustin\footnote{\url{https://data.world/ride-austin/ride-austin-june-6-april-13}} for case study. The data ranges from October 1, 2016 to April 13, 2017, including 1,259,574 trips in total. Similar to the case study in Chicago, we only retained pick-up locations and the corresponding timestamps from the dataset for empirical analysis. All ridesourcing trips were aggregated at the census-tract level on an hourly basis. Finally, the prepared dataset includes 191 census tracts. The first 70\% of the whole dataset was split for model training, followed by the following 10\% for validation and 20\% for testing. Four protected attributes, including the percentage of white, the percentage of low-income households, the percentage of population with a bachelor's degree or above and the percentage of young populations (aged 18-44) were also collected from American Community Survey 2013–2017 5-year estimates data.

\subsection{Models}
\label{benchmark models}
In this study, we applied seven models as the major baseline models to measure the fairness metrics and perform the bias mitigation. We also compared their performance with historical average method. All used models are detailed as follows: 

\begin{itemize}

    \item \textbf{Historical Average} (HA): We calculate the historical average travel demand using the mean values of all observations from the inputted sequence. 
    
    \item \textbf{Multivariate Linear Regression} (MLR): MLR is frequently used in machine learning studies as the benchmark model. This study treats observations at every timestamp $t$ as a covariate.

    \item \textbf{Autoregressive Integrated Moving Average Model} (ARIMA): ARIMA is one of the most fundamental statistical models for forecasting time-series data \cite{makridakis1997arma}. ARIMA consists of three basic parts: auto-regressive, first-differencing and moving-average part. The order of the auto-regressive ($p$) and moving-average ($q$) and the degree of first-differencing ($d$) included should be prespecified before building the model. In this study, we established ARIMA model to predict the travel demand for all areas at once.

    \item \textbf{Multiple Layer Perception} (MLP): MLP is a commonly-used deep neural net model. In this study, the model architecture is set as 1 hidden layer with 300 hidden linear neurons. A drop-out layer rate 0.01 is set after the hidden layer to avoid overfitting. 
    
    \item \textbf{Gated Recurrent Unit} (GRU): GRU is a widely-adopted Recurrent Neural Network (RNN) model with gated hidden neurons \cite{cho2014properties}. GRU can generate the predicted travel demand $x_{i,t+1}$ by inputting the hidden status at timesampe $t-1$ and the travel demand at timestamp $x_{i,t}$. In this way, GRU can dynamically capture the travel demand information at the current timestamp while maintaining the historical demand trend. We use GRU model for forecasting the travel demand for all nodes at once.
    
    \item \textbf{Temporal Graph Convolution Network} (T-GCN): T-GCN can capture the spatial dependency and temporal information at the same time \cite{zhao2019t}. Specifically, the spatial dependency is calibrated by the spatial adjacency graph $G_{adj}$, where 1 indicates two nodes are spatially adjacent and 0 otherwise. T-GCN takes the hidden status at timestamp $t-1$ and the graph-convolution-processed travel demand information at timestamp $t$ as the input. Therefore, T-GCN can effectively deal with data that have strong spatial dependency such traffic speed data.
    
    \item \textbf{Convolutional Long-short Term Memory (ConvLSTM)}: ConvLSTM is one of the most novel approaches for spatio-temporal forecasting problem \cite{shi2015convolutional}. ConvLSTM has a convolution structure in both the input-to-state and state-to-state transitions; it determines a certain cell's future states by considering the inputs and past states from its local neighbors. This characteristic allows it a more powerful strength in handling spatio-temporal correlations. In this study, the convolutional kernel size of the ConvLSTM is set to 5.
    
    \item \textbf{Spatio-Temporal Graph Convolution Network (STGCN)}: STGCN is an effective approach for spatio-temporal traffic flow forecasting \cite{yu2017spatio}. STGCN consists of several spatio-temporal convolution (ST-Conv) blocks. Each block has a ``sandwich''-like structure: two gated sequential convolution layers and one spatial graph convolution layer in between. This allows STGCN to distill the most useful spatial features and capture the most essential temporal features collectively. In this study, we set the number of ST-Conv blocks as 2. Let $d_{i,j}$ denote the distance between node $i$ and node $j$, the element in the weighted adjacency matrix, i.e., $w_{i,j} \in \boldsymbol{W}$, is given by: 
    
    \begin{equation}
    w_{i,j}= \begin{cases}\exp \left(-\frac{d_{i j}^{2}}{\sigma^{2}}\right), & i \neq j \text { and } \exp \left(-\frac{d_{i j}^{2}}{\sigma^{2}}\right) \geq \alpha \\ 0, & \text { otherwise. }\end{cases},
    \end{equation}
    
    \noindent where $\sigma^{2}$ and $\alpha$, assigned as $10^{4}$ and $0.5$, are thresholds that control the sparsity of $\boldsymbol{W}$.
    
\end{itemize}

\section{Results}
\label{results}
This section sequentially reports the modeling results of all benchmark models, the evaluations of their underlying fairness issues and the results after applying our proposed unfairness correction approach. We conducted empirical experiments using the real-world ridesourcing-trip data in Chicago, IL and Austin, TX. The analytical spatial unit is census tract. We incorporate the regularization term into the loss function for all models. All experiments were completed in a Pytorch environment using an Ampere A-100 GPU. We tuned the hyperparameters such as batch size and sequence length under each fairness weight $\lambda$ using grid search. We built our models with Adam optimizer \cite{kingma2014adam}. Early stopping method is also taken to avoid overfitting problems. In this study, we use $60$ and $40$ percentile statistics for the protected attributes as the threshold to determine the label (i.e., $p_j^i$) of each node (e.g., census tract). For instance, the $60$ percentile of white population percentage attribute is $62.35\%$, for nodes with white population percentage over $62.35\%$ are labeled as advantaged.

\subsection{Unfairness detection}

The predictive performance and two fairness metrics (i.e., correlation [Corr] and prediction accuracy gap [PAG]) of all models with respect to four protected attributes are presented in Table \ref{tab:benchmarks_chicago} and Table \ref{tab:benchmarks_austin}.

\begin{table}[!ht]
\centering
\caption{Modeling results of the benchmarks (Chicgao)}
\label{tab:benchmarks_chicago}
\resizebox{\textwidth}{!}{%
\begin{tabular}{@{}lllllllllll@{}}
\toprule
\multirow{2}{*}{Models} & \multirow{2}{*}{MAE} & \multirow{2}{*}{RMSE} & Race &  & Edu &  & Age &  & Income &  \\ \cmidrule(l){4-11} 
         &       &        & Corr   & PAG (\%) & Corr   & PAG (\%) & Corr   & PAG (\%) & Corr   & PAG (\%) \\ \midrule
HA       & 7.703 & 27.630 & 0.058  & -32.452  & 0.072  & -54.277  & 0.048  & -50.220  & -0.031 & -26.303  \\
MLR      & 5.535 & 12.973 & -0.048 & 8.040    & -0.084 & 4.261    & -0.092 & 3.862    & 0.049  & 6.667    \\
ARIMA    & 4.541 & 12.259 & -0.055 & 4.908    & -0.122 & 9.890    & -0.128 & 11.079   & 0.057  & 3.828    \\
MLP      & 3.918 & 10.147 & -0.054 & 4.453    & -0.128 & 9.487    & -0.133 & 10.611   & 0.060  & 3.671    \\
GRU      & 3.715 & 9.069  & -0.058 & 6.293    & -0.143 & 15.161   & -0.128 & 13.432   & 0.071  & 6.367    \\
T-GCN    & 4.705 & 9.993  & -0.093 & 23.672   & -0.148 & 31.644   & -0.153 & 34.748   & 0.095  & 19.175   \\
STGCN    & \textbf{3.012} & 8.539  & -0.122 & 9.679    & -0.279 & 21.363   & -0.285 & 24.484   & 0.129  & 8.182    \\
ConvLSTM & 3.246 & \textbf{8.176}  & -0.075 & 8.958    & -0.144 & 13.063   & -0.150 & 14.696   & 0.104  & 9.392    \\ \bottomrule
\end{tabular}%
}
\parbox[t]{0.98\textwidth}{\vskip3pt{\footnotesize Notes: Corr represents correlation. All correlations are statistically significant at 1\% confidence level.}}
\end{table}

\begin{table}[!ht]
\centering
\caption{Modeling results of the benchmarks (Austin)}
\label{tab:benchmarks_austin}
\resizebox{\textwidth}{!}{%
\begin{tabular}{@{}lllllllllll@{}}
\toprule
\multirow{2}{*}{Models} & \multirow{2}{*}{MAE} & \multirow{2}{*}{RMSE} & Race &  & Edu &  & Age &  & Income &  \\ \cmidrule(l){4-11} 
         &       &       & Corr   & PAG (\%) & Corr   & PAG (\%) & Corr   & PAG (\%) & Corr   & PAG (\%) \\ \midrule
HA       & 1.655 & 8.538 & -0.030 & 5.116    & -0.106 & 6.073    & -0.008 & -8.881   & 0.026  & 6.170    \\
MLR      & 1.324 & 4.280 & -0.008 & 0.572    & -0.041 & 4.118    & -0.056 & 6.561    & -0.029 & -0.784   \\
ARIMA    & 1.335 & 4.695 & -0.008 & 0.316    & -0.048 & 4.940    & -0.078 & 8.578    & -0.047 & -1.758   \\
MLP      & 1.297 & 4.163 & -0.008 & 0.467    & -0.048 & 5.030    & -0.074 & 8.156    & -0.042 & -1.073   \\
GRU      & 1.064 & 3.654 & -0.044 & 2.166    & -0.136 & 10.639   & -0.110 & 10.648   & -0.026 & -0.307   \\
T-GCN    & 1.357 & 3.911 & 0.010  & -1.983   & -0.049 & 5.627    & -0.098 & 12.801   & -0.087 & -6.796   \\
STGCN    & \textbf{1.042} & 4.064 & \textbf{-0.034} & \textbf{1.234}    & -0.178 & 10.933   & -0.179 & 11.614   & \textbf{-0.081} & \textbf{-1.013}   \\
ConvLSTM & 1.057 & \textbf{3.162} & \textbf{0.000}  & \textbf{-0.391}   & -0.088 & 5.289    & -0.080 & 4.642    & \textbf{-0.024} & \textbf{0.888}    \\ \bottomrule
\end{tabular}%
}
\parbox[t]{0.98\textwidth}{\vskip3pt{\footnotesize Notes: Corr represents correlation. All correlations are statistically significant at 1\% confidence level.}}
\end{table}

We show the results of the predictive performance for each benchmark in Chicago ridesourcing-trip data (Table \ref{tab:benchmarks_chicago}) and Austin ridesourcing-trip data (Table \ref{tab:benchmarks_austin}). 

Regarding prediction accuracy, all benchmark models show a similar trend across two case studies. The performance ranking is ConvLSTM $ \approx $ STGCN $>$ GRU $>$ T-GCN $>$ MLP $>$ ARIMA $>$ HA. It indicates that the prediction accuracy gradually increases as the model becomes more complex. Two convolution models, i.e., STGCN and CovLSTM, are best-performing among all models. Both STGCN and ConvLSTM can incorporate spatial and temporal information through the convolution blocks, which enhance their prediction power. Among two RNN-based models, GRU outperformed T-GCN for both MAE and RMSE. MLP, due to its simple model architecture, underperformed all neural network-based models. Compared with deep neural networks, traditional statistical models, i.e., MLR and ARIMA, have relatively low prediction accuracy. However, their performance still significantly outperformed HA. MLR and ARIMA both have a prespecified (linear) model structure and cannot capture the nonlinearity between the inputs and target variables, which restricts the predictive capability. 

Regarding fairness issues, for Chicago ridesourcing-trip data, Table \ref{tab:benchmarks_chicago} shows that HA exhibits completely inverse relationships in correlation and gap compared with other models. Since HA has the worst predictive performance, the corresponding fairness metrics could be unreliable. The results illustrate that both statistical and deep learning models have evident fairness issues. protected attributes, including race, education and age, are negatively correlated with the prediction accuracy which means that communities with high proportion of white population, high education-attainment rate and more young people have high prediction accuracy. Income level is positively related to predictive performance, indicating that communities with more low-income households may have higher perdition error. In terms of magnitude, we found that education and age have the largest value of correlation with prediction accuracy, followed by income and race. Although there are variations in the magnitude of correlations, the signs for all protected attributes among all models except for HA are consistent. In addition to correlations, we also explored the PAG between the advantaged groups and disadvantaged groups. Table \ref{tab:benchmarks_chicago} presents that all gaps have a positive value (except for HA), indicating that the prediction error for minority groups is higher than for advantaged groups. Additionally, the prediction accuracy disparity is more pronounced for education and age than for race and income.


For Austin ridesourcing-trip data, all benchmark models demonstrate a similar performance (both trend and direction of associations) compared with using Chicago dataset. However, results showed that the fairness issues are relatively subdued in Austin dataset. In other words, the extent of unfairness (as shown by correlation coefficient and PAG) is notably diminished in comparison to the Chicago dataset. Notably, Table \ref{tab:benchmarks_austin} shows that prediction accuracy is less biased regarding race and income. Two best-performing models (i.e., STGCN and ConvLSTM) may produce satisfying fair predictions. For example, the correlation between prediction accuracy and race delivered by ConvLSTM is 0.000 and the PAG regarding race is only -0.391\%. This evidence indicates that the unfairness in prediction accuracy should be of little concern for this protected attribute.

\subsection{Unfairness correction}
\label{unfairness_correction}

We tuned a set of values of $\lambda$ (i.e., the weight for fairness loss) by grid search to validate the effectiveness of the proposed unfairness correction method. Table \ref{tab:multiattri_chicago} and Table \ref{tab:multiattri_austin} present the results of simultaneously mitigating the unfairness issues for multiple protected attributes across two case studies. We only present the best $\lambda$ (i.e., the one that can significantly improve fairness while largely preserving prediction accuracy) from the empirical experiments. We also add experimental results of correcting unfairness of only one single attribute at the bottom of each table for comparison. For the sensitivity analysis of $\lambda$, please refer to Section \ref{trade-off_sensitivity}. As discussed in previous section, only very limited prediction accuracy disparities are detected on race (percentage of white population) and income (percentage of low-income households) in the case study of Austin (as shown in Table \ref{tab:benchmarks_austin}). Thus, we decided to only correct the unfairness of prediction accuracy manifested in education (percentage of bachelor holders) and age (percentage of young population) in this case.


\begin{table}[!h]
\centering
\caption{Multi-attribute unfairness correction in Chicago.}
\label{tab:multiattri_chicago}
\resizebox{\textwidth}{!}{%
\begin{tabular}{@{}lllllllll@{}}
\toprule
\multicolumn{9}{l}{\textit{Multi-attribute (debiasing four selected attributes)}}                                                                                                                                                                                                                                                                                                                                                                                                                                             \\ \midrule
Model                   &         & MLR                                                            & ARIMA                                                         & MLP                                                         & GRU                                                         & T-GCN                                                       & STGCN                                                       & ConvLSTM                                                    \\ 
$\lambda$                  &         & 0.025                                                          & 0.025                                                         & 0.025                                                       & 0.05                                                        & 0.075                                                       & 0.025                                                       & 0.025                                                       \\ \midrule
RMSE                    &         & \begin{tabular}[c]{@{}l@{}}12.986\\ (-0.1\%)\end{tabular}      & \begin{tabular}[c]{@{}l@{}}12.308\\ (-0.4\%)\end{tabular}     & \begin{tabular}[c]{@{}l@{}}10.189\\ (-0.414\%)\end{tabular} & \begin{tabular}[c]{@{}l@{}}9.256\\ (-2.062\%)\end{tabular}  & \begin{tabular}[c]{@{}l@{}}10.267\\ (-2.742\%)\end{tabular} & \begin{tabular}[c]{@{}l@{}}8.424\\ (1.347\%)\end{tabular}   & \begin{tabular}[c]{@{}l@{}}8.595\\ (-5.125\%)\end{tabular}  \\ \midrule
\multirow{2}{*}{Race}   & Corr    & \begin{tabular}[c]{@{}l@{}}-0.013\\ (72.917\%)\end{tabular}    & \begin{tabular}[c]{@{}l@{}}-0.005\\ (90.909\%)\end{tabular}   & \begin{tabular}[c]{@{}l@{}}-0.024\\ (55.556\%)\end{tabular} & \begin{tabular}[c]{@{}l@{}}-0.022\\ (62.069\%)\end{tabular} & \begin{tabular}[c]{@{}l@{}}-0.013\\ (86.022\%)\end{tabular} & \begin{tabular}[c]{@{}l@{}}0.007\\ (94.262\%)\end{tabular}  & \begin{tabular}[c]{@{}l@{}}-0.007\\ (90.667\%)\end{tabular} \\ \cmidrule(l){2-9}
                        & PAG(\%) & \begin{tabular}[c]{@{}l@{}}2.279\\ (71.654\%)\end{tabular}     & \begin{tabular}[c]{@{}l@{}}1.176\\ (76.039\%)\end{tabular}    & \begin{tabular}[c]{@{}l@{}}0.913\\ (79.497\%)\end{tabular}  & \begin{tabular}[c]{@{}l@{}}-0.935\\ (85.142\%)\end{tabular} & \begin{tabular}[c]{@{}l@{}}-0.854\\ (96.392\%)\end{tabular} & \begin{tabular}[c]{@{}l@{}}-4.217\\ (56.431\%)\end{tabular} & \begin{tabular}[c]{@{}l@{}}0.649\\ (92.755\%)\end{tabular}  \\ \midrule
\multirow{2}{*}{Edu}    & Corr    & \begin{tabular}[c]{@{}l@{}}-0.009\\ (89.286\%)\end{tabular}    & \begin{tabular}[c]{@{}l@{}}-0.011\\ (90.984\%)\end{tabular}   & \begin{tabular}[c]{@{}l@{}}-0.03\\ (76.563\%)\end{tabular}  & \begin{tabular}[c]{@{}l@{}}-0.031\\ (78.322\%)\end{tabular} & \begin{tabular}[c]{@{}l@{}}-0.029\\ (80.405\%)\end{tabular} & \begin{tabular}[c]{@{}l@{}}0.007\\ (97.491\%)\end{tabular}  & \begin{tabular}[c]{@{}l@{}}-0.011\\ (92.361\%)\end{tabular} \\ \cmidrule(l){2-9}
                        & PAG(\%) & \begin{tabular}[c]{@{}l@{}}-16.956\\ (-297.935\%)\end{tabular} & \begin{tabular}[c]{@{}l@{}}-13.507\\ (-36.572\%)\end{tabular} & \begin{tabular}[c]{@{}l@{}}-2.874\\ (69.706\%)\end{tabular} & \begin{tabular}[c]{@{}l@{}}-2.064\\ (86.386\%)\end{tabular} & \begin{tabular}[c]{@{}l@{}}-3.979\\ (87.426\%)\end{tabular} & \begin{tabular}[c]{@{}l@{}}-12.47\\ (41.628\%)\end{tabular} & \begin{tabular}[c]{@{}l@{}}-6.612\\ (49.384\%)\end{tabular} \\ \midrule
\multirow{2}{*}{Age}    & Corr    & \begin{tabular}[c]{@{}l@{}}-0.008\\ (91.304\%)\end{tabular}    & \begin{tabular}[c]{@{}l@{}}-0.011\\ (91.406\%)\end{tabular}   & \begin{tabular}[c]{@{}l@{}}-0.035\\ (73.684\%)\end{tabular} & \begin{tabular}[c]{@{}l@{}}-0.039\\ (69.531\%)\end{tabular} & \begin{tabular}[c]{@{}l@{}}-0.033\\ (78.431\%)\end{tabular} & \begin{tabular}[c]{@{}l@{}}0.01\\ (96.491\%)\end{tabular}   & \begin{tabular}[c]{@{}l@{}}-0.009\\ (94\%)\end{tabular}     \\ \cmidrule(l){2-9}
                        & PAG(\%) & \begin{tabular}[c]{@{}l@{}}-14.327\\ (-270.974\%)\end{tabular} & \begin{tabular}[c]{@{}l@{}}-5.291\\ (52.243\%)\end{tabular}   & \begin{tabular}[c]{@{}l@{}}-3.926\\ (63.001\%)\end{tabular} & \begin{tabular}[c]{@{}l@{}}-1.074\\ (92.004\%)\end{tabular} & \begin{tabular}[c]{@{}l@{}}2.623\\ (92.451\%)\end{tabular}  & \begin{tabular}[c]{@{}l@{}}-6.386\\ (73.918\%)\end{tabular} & \begin{tabular}[c]{@{}l@{}}-6.441\\ (56.172\%)\end{tabular} \\ \midrule
\multirow{2}{*}{Income} & Corr    & \begin{tabular}[c]{@{}l@{}}0.016\\ (67.347\%)\end{tabular}     & \begin{tabular}[c]{@{}l@{}}0.008\\ (85.965\%)\end{tabular}    & \begin{tabular}[c]{@{}l@{}}0.017\\ (71.667\%)\end{tabular}  & \begin{tabular}[c]{@{}l@{}}0.015\\ (78.873\%)\end{tabular}  & \begin{tabular}[c]{@{}l@{}}0.011\\ (88.421\%)\end{tabular}  & \begin{tabular}[c]{@{}l@{}}-0.003\\ (97.674\%)\end{tabular} & \begin{tabular}[c]{@{}l@{}}0.013\\ (87.5\%)\end{tabular}    \\ \cmidrule(l){2-9}
                        & PAG(\%) & \begin{tabular}[c]{@{}l@{}}-1.527\\ (77.096\%)\end{tabular}    & \begin{tabular}[c]{@{}l@{}}-3.521\\ (8.02\%)\end{tabular}     & \begin{tabular}[c]{@{}l@{}}-0.312\\ (91.501\%)\end{tabular} & \begin{tabular}[c]{@{}l@{}}-0.323\\ (94.927\%)\end{tabular} & \begin{tabular}[c]{@{}l@{}}-0.403\\ (97.898\%)\end{tabular} & \begin{tabular}[c]{@{}l@{}}-3.688\\ (54.925\%)\end{tabular} & \begin{tabular}[c]{@{}l@{}}0.462\\ (95.081\%)\end{tabular}  \\ \midrule
\multicolumn{9}{l}{\textit{Single-attribute (only debiasing Income)}}                                                                                                                                                                                                                                                                                                                                                                                                                    \\ \midrule
\multirow{4}{*}{Income} & $\lambda$  & 0.1                                                            & 0.025                                                         & 0.025                                                       & 0.025                                                       & 0.025                                                       & 0.2                                                         & 0.025                                                       \\ \cmidrule(l){2-9}
                        & RMSE    & \begin{tabular}[c]{@{}l@{}}13.005\\ (-0.247\%)\end{tabular}    & \begin{tabular}[c]{@{}l@{}}12.263\\ (-0.033\%)\end{tabular}   & \begin{tabular}[c]{@{}l@{}}10.209\\ (-0.611\%)\end{tabular} & \begin{tabular}[c]{@{}l@{}}9.32\\ (-2.768\%)\end{tabular}   & \begin{tabular}[c]{@{}l@{}}10.641\\ (-6.485\%)\end{tabular} & \begin{tabular}[c]{@{}l@{}}9.029\\ (-5.738\%)\end{tabular}  & \begin{tabular}[c]{@{}l@{}}8.307\\ (-1.602\%)\end{tabular}  \\ \cmidrule(l){2-9}
                        & Corr    & \begin{tabular}[c]{@{}l@{}}0.015\\ (69.388\%)\end{tabular}     & \begin{tabular}[c]{@{}l@{}}0\\ (100\%)\end{tabular}           & \begin{tabular}[c]{@{}l@{}}0.015\\ (75\%)\end{tabular}      & \begin{tabular}[c]{@{}l@{}}0.003\\ (95.775\%)\end{tabular}  & \begin{tabular}[c]{@{}l@{}}-0.008\\ (91.579\%)\end{tabular} & \begin{tabular}[c]{@{}l@{}}-0.001\\ (99.225\%)\end{tabular} & \begin{tabular}[c]{@{}l@{}}0.002\\ (98.077\%)\end{tabular}  \\ \cmidrule(l){2-9}
                        & PAG(\%) & \begin{tabular}[c]{@{}l@{}}-0.03\\ (99.55\%)\end{tabular}      & \begin{tabular}[c]{@{}l@{}}-3.566\\ (6.844\%)\end{tabular}    & \begin{tabular}[c]{@{}l@{}}-0.521\\ (85.808\%)\end{tabular} & \begin{tabular}[c]{@{}l@{}}-2.204\\ (65.384\%)\end{tabular} & \begin{tabular}[c]{@{}l@{}}-5.303\\ (72.344\%)\end{tabular} & \begin{tabular}[c]{@{}l@{}}2.57\\ (68.59\%)\end{tabular}    & \begin{tabular}[c]{@{}l@{}}-3.695\\ (60.658\%)\end{tabular} \\ \bottomrule
\end{tabular}%
}

\parbox[t]{0.98\textwidth}{\vskip3pt{\footnotesize Notes: Corr represents correlation. PAG refers to prediction accuracy gap. The value inside each bracket refers to the percentage change of metric in absolute value. It is computed as: $(|o|-|m|)*100\%/|o|$, with $o$ denoting the initial value obtained from the fairness-unaware model and $m$ representing the final value from the fairness-aware model. A positive value indicates the improvement while a negative value indicates the reduction.}}

\end{table}


\begin{table}[!h]
\centering
\caption{Multi-attribute unfairness correction in Austin. }
\label{tab:multiattri_austin}
\resizebox{\textwidth}{!}{%
\begin{tabular}{@{}lllllllll@{}}
\toprule
\multicolumn{9}{l}{\textit{Multi-attribute (debiasing two selected attributes)}}                                                                                                                                                                                                                                                                                                                                                                                                                                  \\ \midrule
Model                &         & MLR                                                         & ARIMA                                                       & MLP                                                         & GRU                                                         & T-GCN                                                       & STGCN                                                       & ConvLSTM                                                    \\ 
$\lambda$               &         & 0.025                                                       & 0.025                                                       & 0.5                                                         & 0.05                                                        & 0.4                                                         & 0.05                                                        & 0.025                                                       \\ \midrule
RMSE                 &         & \begin{tabular}[c]{@{}l@{}}4.28\\ (0.005\%)\end{tabular}    & \begin{tabular}[c]{@{}l@{}}4.697\\ (-0.037\%)\end{tabular}  & \begin{tabular}[c]{@{}l@{}}4.158\\ (0.128\%)\end{tabular}   & \begin{tabular}[c]{@{}l@{}}3.983\\ (-9.000\%)\end{tabular}  & \begin{tabular}[c]{@{}l@{}}3.694\\ (5.541\%)\end{tabular}   & \begin{tabular}[c]{@{}l@{}}4.691\\ (-15.433\%)\end{tabular} & \begin{tabular}[c]{@{}l@{}}3.341\\ (-5.651\%)\end{tabular}  \\ \midrule
\multirow{2}{*}{Edu} & Corr    & \begin{tabular}[c]{@{}l@{}}-0.009\\ (78.071\%)\end{tabular} & \begin{tabular}[c]{@{}l@{}}-0.01\\ (79.285\%)\end{tabular}  & \begin{tabular}[c]{@{}l@{}}-0.004\\ (91.677\%)\end{tabular} & \begin{tabular}[c]{@{}l@{}}-0.007\\ (94.849\%)\end{tabular} & \begin{tabular}[c]{@{}l@{}}0.002\\ (95.877\%)\end{tabular}  & \begin{tabular}[c]{@{}l@{}}-0.002\\ (98.874\%)\end{tabular} & \begin{tabular}[c]{@{}l@{}}0.01\\ (88.668\%)\end{tabular}   \\ \cmidrule(l){2-9}
                     & PAG(\%) & \begin{tabular}[c]{@{}l@{}}1.938\\ (52.941\%)\end{tabular}  & \begin{tabular}[c]{@{}l@{}}2.41\\ (51.213\%)\end{tabular}   & \begin{tabular}[c]{@{}l@{}}-0.171\\ (96.6\%)\end{tabular}   & \begin{tabular}[c]{@{}l@{}}-0.365\\ (96.569\%)\end{tabular} & \begin{tabular}[c]{@{}l@{}}2.038\\ (63.783\%)\end{tabular}  & \begin{tabular}[c]{@{}l@{}}1.401\\ (87.186\%)\end{tabular}  & \begin{tabular}[c]{@{}l@{}}-1.641\\ (68.973\%)\end{tabular} \\ \midrule
\multirow{2}{*}{Age} & Corr    & \begin{tabular}[c]{@{}l@{}}-0.01\\ (82.214\%)\end{tabular}  & \begin{tabular}[c]{@{}l@{}}-0.014\\ (81.956\%)\end{tabular} & \begin{tabular}[c]{@{}l@{}}-0.015\\ (79.841\%)\end{tabular} & \begin{tabular}[c]{@{}l@{}}-0.028\\ (74.578\%)\end{tabular} & \begin{tabular}[c]{@{}l@{}}0.000\\ (100\%)\end{tabular}     & \begin{tabular}[c]{@{}l@{}}0.025\\ (86.027\%)\end{tabular}  & \begin{tabular}[c]{@{}l@{}}-0.006\\ (92.489\%)\end{tabular} \\ \cmidrule(l){2-9}
                     & PAG(\%) & \begin{tabular}[c]{@{}l@{}}2.342\\ (64.305\%)\end{tabular}  & \begin{tabular}[c]{@{}l@{}}3.458\\ (59.685\%)\end{tabular}  & \begin{tabular}[c]{@{}l@{}}0.833\\ (89.786\%)\end{tabular}  & \begin{tabular}[c]{@{}l@{}}4.180\\ (60.745\%)\end{tabular}  & \begin{tabular}[c]{@{}l@{}}1.045\\ (91.836\%)\end{tabular}  & \begin{tabular}[c]{@{}l@{}}-1.269\\ (89.073\%)\end{tabular} & \begin{tabular}[c]{@{}l@{}}-0.534\\ (88.495\%)\end{tabular} \\ \midrule
\multicolumn{9}{l}{\textit{Single-attribute (only debiasing Age)}}                                                                                                                                                                                                                                                                                                                                                                                                               \\ \midrule
\multirow{4}{*}{Age} & $\lambda$  & 0.075                                                       & 0.05                                                        & 0.1                                                         & 0.025                                                       & 0.025                                                       & 0.05                                                        & 0.025                                                       \\ \cmidrule(l){2-9}
                     & RMSE    & \begin{tabular}[c]{@{}l@{}}4.281\\ (-0.019\%)\end{tabular}  & \begin{tabular}[c]{@{}l@{}}4.703\\ (-0.165\%)\end{tabular}  & \begin{tabular}[c]{@{}l@{}}4.172\\ (-0.208\%)\end{tabular}  & \begin{tabular}[c]{@{}l@{}}3.517\\ (3.752\%)\end{tabular}   & \begin{tabular}[c]{@{}l@{}}3.692\\ (5.593\%)\end{tabular}   & \begin{tabular}[c]{@{}l@{}}4.392\\ (-8.075\%)\end{tabular}  & \begin{tabular}[c]{@{}l@{}}3.354\\ (-6.062\%)\end{tabular}  \\ \cmidrule(l){2-9}
                     & Corr    & \begin{tabular}[c]{@{}l@{}}-0.006\\ (89.122\%)\end{tabular} & \begin{tabular}[c]{@{}l@{}}0.000\\ (99.553\%)\end{tabular}  & \begin{tabular}[c]{@{}l@{}}-0.003\\ (96.292\%)\end{tabular} & \begin{tabular}[c]{@{}l@{}}-0.003\\ (96.292\%)\end{tabular} & \begin{tabular}[c]{@{}l@{}}-0.014\\ (85.273\%)\end{tabular} & \begin{tabular}[c]{@{}l@{}}-0.009\\ (94.822\%)\end{tabular} & \begin{tabular}[c]{@{}l@{}}0.000\\ (99.559\%)\end{tabular}  \\ \cmidrule(l){2-9}
                     & PAG(\%) & \begin{tabular}[c]{@{}l@{}}0.839\\ (87.212\%)\end{tabular}  & \begin{tabular}[c]{@{}l@{}}2.961\\ (65.48\%)\end{tabular}   & \begin{tabular}[c]{@{}l@{}}-0.081\\ (99.007\%)\end{tabular} & \begin{tabular}[c]{@{}l@{}}0.923\\ (91.332\%)\end{tabular}  & \begin{tabular}[c]{@{}l@{}}5.788\\ (54.783\%)\end{tabular}  & \begin{tabular}[c]{@{}l@{}}1.331\\ (88.539\%)\end{tabular}  & \begin{tabular}[c]{@{}l@{}}0.771\\ (83.389\%)\end{tabular}  \\ \bottomrule
\end{tabular}%
}
\parbox[t]{0.98\textwidth}{\vskip3pt{\footnotesize Notes: Corr represents correlation. PAG refers to prediction accuracy gap. The value inside each bracket refers to the percentage change of metric in absolute value. It is computed as: $(|o|-|m|)*100\%/|o|$, with $o$ denoting the initial value obtained from the fairness-unaware model and $m$ representing the final value from the fairness-aware model. A positive value indicates the improvement while a negative value indicates the reduction.}}
\end{table}


There are several key findings to highlight. First, results of the multi-attribute scenario show great consistency across two datasets. Table \ref{tab:multiattri_chicago} and Table \ref{tab:multiattri_austin} show that in almost all trails, incorporating a small fairness weight can significantly reduce the absolute value of the correlation and PAG across all protected attributes. For example, in Chicago dataset, incorporating only 0.050 fairness weight for GRU can lead to 85.142\%, 86.386\%, 92.004\%, 94.927\% reduction of the absolute values of the PAG for race, education, age and income, respectively. In the meantime, the correlation between prediction accuracy and protected attributes also improved more than 60\%, but RMSE only increased by 2.062\%. In Austin dataset, setting $\lambda$ as 0.025 for ConvLSTM yields 68.973\% and 88.496\% PAG shrinkage on education and age by sacrificing only 5.661\% increase on RMSE (from 3.162 to 3.341).

Second, the effects of the proposed unfariness correction method varies across models and protected attributes. For example, Table \ref{tab:multiattri_chicago} shows that when mitigating the income bias, setting $\lambda$ as 0.025 only reduces 54.923\% of the PAG in absolute value for STGCN; while for ConvLSTM, the same setting can lead to a 95.081\% reduction. In addition, the case study on Chicago ridesourcing-trip data reveals that compared with education and age, the absolute value of PAG for race and income are more likely to be largely reduced (i.e., to less than 1\%).

Third, by choosing an appropriate $\lambda$, both fairness and accuracy can be improved at the same time. Taking Chicago dataset as an example, adding 0.025 fairness weight on STGCN can simultaneously reduce the absolute value of PAG and correlations for all protected attributes while even reducing RMSE by 1.347\%.

Moreover, we found that MLR and ARIMA showed limited capabilities in mitigating unfairness. In Chicago dataset, the prediction accuracy disparities of education and age (as shown in the change of PAG) for MLR and ARIMA even increased after debiasing multiple protected attributes. Also, our examination of the Austin dataset indicated that after incorporating the proposed fairness regularization term, although the PAG for MLR and ARIMA decreased, the magnitude of this reduction was comparatively modest in comparison to other models. In fact, these two models are less flexible compared with other deep learning models since they have a pre-specified model structure. We believe that this inherent limitation could hinder their effectiveness in addressing fairness concerns.

Lastly, in most cases, our proposed multi-attribute unfairness correction method shows better performance in reducing disparities of prediction and preserving fairness compared with only debiasing a single attribute, especially for more complex deep learning models (e.g., GRU, T-GCN, STGCN and ConvLSTM). For example, Table \ref{tab:multiattri_chicago} shows that when considering multiple attributes together, GRU and ConvLSTM can close more than 94\% of PAG of income in absolute value; while for the single-attribute scenario, the PAG is only reduced by around 60\%. However, we also observed in certain cases, single-attribute unfairness correction could produce fairer performance. For example, GRU is found to be more effective in reducing PAG when only debiasing age for Austin dataset.


\begin{table}[!h]
\centering
\caption{Performance comparison between only debiasing Age and simultaneously debiasing multiple attributes using ConvLSTM.}
\label{tab:gap_compare_single_multi}
\resizebox{0.9\textwidth}{!}{%
\begin{tabular}{@{}lllllll@{}}
\toprule
\multirow{2}{*}{Model}               & \multirow{2}{*}{$\lambda$} & \multirow{2}{*}{RMSE} & \multicolumn{4}{l}{PAG (\%)}                        \\ \cmidrule(l){4-7} 
                                     &                         &                       & Race            & Edu    & Age    & Income          \\ \midrule
\textit{Chicago}                     &                         &                       &                 &        &        &                 \\
ConvLSTM (Original)                  & 0.000                   & 8.176                 & \textbf{8.958}           & 13.063 & 14.696 & \textbf{9.392}           \\
ConvLSTM (debiasing only Age)        & 0.050                   & 8.200                 & \textbf{10.520} & -2.643 & -7.673 & \textbf{12.159} \\
ConvLSTM (debiasing multi-attribute) & 0.025                   & 8.595                 & 0.649           & -6.612 & -6.441 & 0.462           \\ \midrule
\textit{Austin}                      &                         &                       &                 &        &        &                 \\
ConvLSTM (Original)                  & 0.000                   & 3.162                 & \textbf{-0.391}          & 5.289  & 4.642  & \textbf{0.888}           \\
ConvLSTM (debiasing only Age)        & 0.050                   & 3.354                 & \textbf{2.027}  & 3.993  & 0.771  & \textbf{8.238}  \\
ConvLSTM (debiasing multi-attribute) & 0.025                   & 3.341                 & 1.558           & -1.641 & -0.534 & 0.295           \\ \bottomrule
\end{tabular}%
}
\end{table}

To provide a more comprehensive demonstration of the efficacy of the proposed multi-attribute unfairness correction approach and to pinpoint potential shortcomings in the single-attribute bias correction method, we conduct a comparative analysis of unfairness correction outcomes achieved through debiasing the age variable alone versus debiasing multiple attributes simultaneously. We have chosen the top-performing model, i.e., ConvLSTM, for demonstration. The resulting findings can be found in Table \ref{tab:gap_compare_single_multi}. 

We found that correcting unfairness regarding one attribute might even create more biases for other protected attributes, which aligns with one previous study \citep{zheng2021equality}. This finding highlights the importance of considering multiple protected attributes at once. Specifically, results showed that compared with the original model that purely focused on prediction accuracy, solely correcting unfairness of age variable could indeed help drop the absolute value of PAG. However, by only considering age, the PAG for other variables, especially for race and income variables, even increases. For example, in Austin dataset, debiasing only Age shrank the PAG from 4.642\% to 0.771\% by significantly sacrificing the PAG of income from 0.888\% to 8.238\%. This unexpected outcome may further shed light on the fact that the transportation resource allocations intended to be fair for distinct age groups could nonetheless still be unfair regarding communities with different income levels. Notably, the results showed that the proposed multi-attribute unfairness correction method can effectively debias multiple protected attributes and in almost all cases the absolute value of PAG is significantly dropped compared with the original model without sacrificing too much prediction accuracy.

\subsection{Sensitivity analysis of fairness weight}
\label{trade-off_sensitivity}

We also explored the influence of the fairness weight, i.e., $\lambda$, in shaping the interaction between accuracy and fairness based on the predictive performance of seven models with four protected attributes. Fig. \ref{fig:sensitivity_analysis_final} presented in \ref{appendix_c} illustrates the sensitivity analysis of $\lambda$ in determining accuracy and fairness. The $x$-axis is the value of $\lambda$ while the $y$-axis is the performance metrics (RMSE, correlation coefficient and PAG). Generally, the accuracy for all models decreases when $\lambda$ gradually increases. In terms of RMSE, the marginal effect of $\lambda$ on more complex models is relatively small. Figures show that as $\lambda$ grows, the correlation will first drastically increase/decrease, and then remain flat. Notably, setting a small weight ($\lambda \leq 0.1$) can lead the correlation drop to around 0. The PAG shows a decreasing trend as $\lambda$ gradually increases. But in most cases, the gap may get over-corrected when $\lambda$ is greater than 0.1. According to the tables shown in Section \ref{unfairness_correction}, a suitable fairness weight possibly exists in the range between 0 to 0.1. This finding further reinforces the effectiveness of our proposed unfairness correction approach: incorporating only a small amount of weight for fairness can lead to a significant improvement in producing fair predictions. We also found that increasing fairness weight may not monotonically reduce the PAG. This finding echoes the results in \citet{zheng2021equality}, where they showed that increasing fairness weight might even extend the PAG. Our computational experiments show that this scenario frequently occurs for traditional statistical models. This finding also suggests the need for more fine-grained searching ranges of $\lambda$ when conducting hyperparameter tuning. Overall, the sensitivity of the effects of $\lambda$ shows great consistency across two case studies. Finally, we noticed that in Austin case, setting fairness weight as 0.4 for GRU led a substantial increase in RMSE and PAG. One possible reason could be that this combination of hyperparameters might explode the gradients and thus lead to this numerical instability.

\subsection{Comparison with benchmark fairness regularization methods}

\begin{table}[!h]
\centering
\caption{Comparison with state-of-the-art benchmark regularizers}
\label{tab:compare_benchmark}
\resizebox{\textwidth}{!}{%
\begin{tabular}{@{}lllllllllll@{}}
\toprule
\multirow{2}{*}{Model} & \multirow{2}{*}{Regularizer} & \multicolumn{4}{l}{Chicago (Race)}                           & \multirow{2}{*}{Regularizer} & \multicolumn{4}{l}{Austin (Education)}                            \\ \cmidrule(lr){3-6} \cmidrule(l){8-11} 
                       &                              & Lambda & RMSE            & Corr            & PAG(\%)         &                              & Lambda & RMSE           & Corr            & PAG(\%)         \\ \midrule
MLR                    & $r(e_t,z_j)$                            & 0.025  & \textbf{12.975} & \textbf{-0.001} & \textbf{-1.980} & $r(e_t,z_j)$                            & 0.025  & 4.281 & -0.015 & \textbf{-0.318} \\
                       & EM                           & 0.025  & 15.854          & -0.124          & 60.016          & EM                           & 0.025  & \textbf{4.270}          & -0.030          & 2.829           \\
                       & RFG                          & 0.025  & 13.100          & -0.002          & -3.355          & RFG                          & 0.025  & 4.274          & -0.055          & 5.609           \\
                       & IFG                          & 0.025  & 13.048          & 0.003           & -4.619          & IFG                          & 0.025  & 4.279          & \textbf{-0.005}          & 0.621           \\
ARIMA                  & $r(e_t,z_j)$                            & 0.025  & \textbf{12.263} & -0.021 & \textbf{0.424}  & $r(e_t,z_j)$                            & 0.025  & 4.701 & \textbf{0.005}  & \textbf{0.071}  \\
                       & EM                           & 0.025  & 15.679          & -0.136          & 62.380          & EM                           & 0.025  & 4.702          & -0.047          & 4.730           \\
                       & RFG                          & 0.025  & 12.433          & \textbf{0.000}           & -4.053          & RFG                          & 0.025  & 4.698          & -0.071          & 7.268           \\
                       & IFG                          & 0.025  & 12.375          & 0.009           & -7.000          & IFG                          & 0.025  & \textbf{4.696}          & -0.013          & 1.776           \\
MLP                    & $r(e_t,z_j)$                            & 0.050  & \textbf{10.142} & -0.022 & \textbf{0.059}  & $r(e_t,z_j)$                            & 0.050  & 4.166 & \textbf{-0.012} & \textbf{0.774}  \\
                       & EM                           & 0.050  & 13.119          & \textbf{-0.014}          & 0.316           & EM                           & 0.050  & 4.120          & -0.050          & 4.898           \\
                       & RFG                          & 0.050  & 10.259          & -0.026          & 0.483           & RFG                          & 0.050  & \textbf{4.110}          & -0.045          & 4.261           \\
                       & IFG                          & 0.050  & 10.221          & -0.034          & 2.027           & IFG                          & 0.050  & 4.147          & -0.039          & 3.727           \\
GRU                    & $r(e_t,z_j)$                            & 0.025  & \textbf{9.263}  & \textbf{-0.002} & -2.940 & $r(e_t,z_j)$                            & 0.025  & \textbf{3.517} & \textbf{-0.003} & \textbf{0.923}  \\
                       & EM                           & 0.025  & 14.455          & -0.141          & 120.286         & EM                           & 0.025  & 4.329          & -0.111          & 9.201           \\
                       & RFG                          & 0.025  & 10.109          & -0.017          & \textbf{0.714}           & RFG                          & 0.025  & 3.545          & -0.051          & 3.926           \\
                       & IFG                          & 0.025  & 9.325           & -0.072          & 10.440          & IFG                          & 0.025  & 3.564          & -0.060          & 4.744           \\
T-GCN                  & $r(e_t,z_j)$                            & 0.075  & \textbf{10.103} & \textbf{0.003}  & \textbf{-3.958} & $r(e_t,z_j)$                            & 0.200  & \textbf{3.767} & \textbf{-0.001} & \textbf{0.898}  \\
                       & EM                           & 0.075  & 16.722          & -0.108          & 102.303         & EM                           & 0.200  & 4.050          & -0.048          & 7.186           \\
                       & RFG                          & 0.075  & 12.215          & -0.034          & 17.875          & RFG                          & 0.200  & 3.990          & -0.018          & 3.570           \\
                       & IFG                          & 0.075  & 12.621          & -0.074          & 31.545          & IFG                          & 0.200  & 3.982          & 0.005           & 2.119           \\
STGCN                  & $r(e_t,z_j)$                            & 0.075  & \textbf{8.790}  & 0.005  & \textbf{-1.486} & $r(e_t,z_j)$                            & 0.300  & \textbf{4.062} & 0.002  & \textbf{0.835}  \\
                       & EM                           & 0.075  & 12.891          & -0.009          & 45.571          & EM                           & 0.300  & 4.576          & -0.059          & 6.618           \\
                       & RFG                          & 0.075  & 10.216          & 0.009           & 15.865          & RFG                          & 0.300  & 4.092          & \textbf{0.000}           & -2.710          \\
                       & IFG                          & 0.075  & 9.904           & \textbf{-0.002}          & 5.433           & IFG                          & 0.300  & 4.403          & -0.018          & 3.307           \\
ConvLSTM               & $r(e_t,z_j)$                            & 0.025  & 8.474  & \textbf{-0.001} & \textbf{-0.389} & $r(e_t,z_j)$                            & 0.025  & \textbf{3.325} & \textbf{-0.001} & \textbf{0.497}  \\
                       & EM                           & 0.025  & 12.634          & -0.153          & 92.676          & EM                           & 0.025  & 3.474          & -0.097          & 6.164           \\
                       & RFG                          & 0.025  & 8.711           & -0.035          & 7.152           & RFG                          & 0.025  & 3.437          & -0.146          & 9.645           \\
                       & IFG                          & 0.025  & \textbf{8.226}           & -0.083          & 13.887          & IFG                          & 0.025  & 3.382          & -0.069          & 4.705           \\ \bottomrule
\end{tabular}%
}
\end{table}

This study compares the performance of the proposed unfairness correction approach (i.e., the absolute correlation regularization term) with three state-of-the-art benchmark regularizers, including Equal Mean (EM) \citep{calders2013controlling}, Region-based Fairness Gap (RFG) and Individual-based Fairness Gap (IFG) \citep{yan2020fairness}. For experiments, we only consider single-attribute scenario as these three benchmark regularizers are explicitly designed for addressing unfairness of a single protected attribute. For Chicago Ridesourcing dataset, we select race (percentage of white population) for model debiasing; while for RideAustin dataset, education variable (percentage of bachelor holders) is chosen for comparison. All benchmark regularizers are set with the best-performing $\lambda$ yielded by our proposed method for comparison.

Table \ref{tab:compare_benchmark} presents the comparative analysis between our proposed method (i.e., absolute correlation regularizer) and three state-of-the-art benchmark regularizers. Results unequivocally show that the proposed method evidently outperforms other methods in terms of preserving prediction accuracy as well as improving fairness. Among all regularizers, EM delivers the worst performance. This is expected since EM focuses on balancing the target variable (i.e., ridesourcing demand) of disadvantaged and advantaged groups instead of the prediction accuracy. However, this method could be questionable since the variations in ridesourcing usage between different population groups may naturally exist due to socioeconomic and demographic disparities \citep{brown2019prevalence}. RFG and IFG tend to yield improved outcomes in terms of both accuracy and fairness when compared to EM. Moreover, in certain scenarios, their performance (especially for correlation and RMSE) surpasses that of the proposed method. We attribute this to their capabilities to effectively reduce variations in per capita travel demand for each individual population group, as indicated in \citet{yan2020fairness}. However, these two metrics may still not be able to fully account for the inherent disparities of different population groups in generating travel demand \citep{zhang2022machine}. In most cases, especially for deep learning models with more complex model architectures, the proposed method can significantly help reduce the PAG between disadvantaged and advantaged groups while keeping the prediction error lowest. Although in some cases the proposed method may not be the best-performing one regarding accuracy, the produced RMSE still remains within an acceptable range.

\section{Discussion}
\label{discussion}
The above sections demonstrate the modeling results of our proposed unfairness correction method. In this section, we will discuss the merits of the unfairness correction method, policy implications, and the limitations of the work and future research directions.

\subsection{Merit of the unfairness correction method}

The merit of the proposed unfairness correction method is threefold.

First, \textit{a new regularizer to simultaneously debias multiple protected attributes}. The current literature rarely discusses how to effectively address fairness issues for multiple protected attributes. However, designing a method that can accommodate various fairness needs is necessary for real-world applications \citep{wan2023processing}. This study addresses this issue by proposing to use \textbf{Multiple Correlation Coefficient} (i.e., $R$ of a linear model) as a regularization term and incorporating it into the loss function. The multiple correlation coefficient can directly measure the correlation between the target variable (i.e., prediction accuracy) and a set of protected demographic variables (i.e., race, age, education and income). By minimizing the coefficient of multiple correlation, AI models can simultaneously debias multiple sensitive attributes. Unlike adding multiple regularization terms (one for each attribute) \citep{yan2020fairness}, this approach is straightforward and easy to implement, and there is no need to fine-tune the fairness weight for different attributes (only one is enough). Also, this approach has little concern about the \textit{multicolinearity} (as shown in Appendix.\ref{appendix_b}) issues among different protected attributes, since the goal of the linear model is to use the set of protected attributes to forecast the prediction errors instead of estimating and interpreting the beta coefficients \citep{shmueli2010explain}. Overall, our proposed unfairness correction method enables future studies to flexibly debias both single or multiple protected attributes of interests.

Second, \textit{flexibility and transparency}. The proposed unfairness correction method is model-agnostic and may be generalizable for different applications and different data modalities. We implemented the unfairness correction method on both statistical and deep learning models. Results jointly demonstrated that, generally, this approach could mitigate the unfairness while only slightly reducing the overall accuracy. Specifically, we correct the unfairness by incorporating an explicitly designed absolute correlation regularization term into the loss function without modifying the model structure. It allows the unfairness correction method great flexibility to be independent of the underlying model. Scholars can thus flexibly adopt any model they want in addressing fairness issues. Also, the proposed method enjoys great transparency since end-users (e.g., stakeholders) can easily understand how fairness is being taken into account and improved (from the fairness regularization term). Moreover, this method is transferable for other forecasting applications. Besides travel demand forecasting, other important issues including traffic count forecasting, pedestrian activity forecasting or crash frequency forecasting may also have silent fairness problems. Researchers can apply our proposed method to address the fairness issues and provide fair decision-making. This study only examined the proposed method using time-series (panel) data. However, we believe it can be easily generalized to other applications with different data modality. For example, transportation-planning models, which usually use cross-sectional data, should also be examined with fairness analysis. Our unfairness correction method can be flexibly adopted by planning models \citep[e.g.][]{zhang2022machine} to inform fair design of transportation ecosystems. Flexibility is also reflected in that, once the models are trained, access to protected attributes is no longer required. Unlike the post-processing technique that always requires access to the protected attribute \citep{hardt2016equality, agarwal2019fair}, our approach lifts this restriction and can be flexibly adapted for future forecasting tasks.

Third, \textit{effectiveness in achieving fairness while preserving prediction accuracy}. Multiple studies reported that machine learning has a trade-off between accuracy and fairness (e.g., \citep{berk2017convex, agarwal2019fair}), i.e., the reduction of unfairness will inevitably trigger an accuracy drop. Our scheme addresses this trade-off by incorporating an interactive weight coefficient (i.e., $\lambda$) into the loss function. We treat $\lambda$ as a hyperparameter of the learning task and tune it together with other hyperparameters. In this way, the model automatically finds the optimal hyperparameter combination that has the best performance in improving fairness while maintaining prediction accuracy. Most of our experiments revealed that this approach could significantly reduce unfairness only at little expense of accuracy decline. While in some cases, our proposed method can even significantly improve fairness and slightly improve prediction accuracy.


\subsection{Policy implications}


Dynamically balancing the supply and demand for transportation systems is important to improve cost-benefit effects and efficiency. And this balance relies heavily on accurate predictions \citep{chu2019deep}. Although machine learning intensively promotes predictions, it may simultaneously introduce bias. The overall satisfactory predictions may hide a huge prediction accuracy gap across areas of the city or underrepresented groups of residents \citep{yan2020fairness, zheng2021equality}. Our study also confirms this finding. Specifically, Table \ref{tab:benchmarks_chicago} shows that both machine learning and statistical models can produce lower prediction accuracy for the disadvantaged communities (i.e., the non-white-majority, the lower-education-attainment, the elderly and the low-income) than that of the advantaged communities. The predictive disparity implies that if transportation planners naively use such travel demand forecasting models without accounting for the fairness issues, the modeling results will lead to ineffective transportation resource allocations, impede the mobility of the disadvantaged communities, and even possibly further exacerbate the existing operational biases of ridesourcing services, e.g., higher trip-cancellation rate, longer waiting times and higher per-mile fees for disadvantaged communities \citep{brown2019equalizer, yang2021equitable, brown2022not, brown2021equity}.

Our proposed method can help mitigate the unfairness issues of the current ridesourcing operations to better serve the disadvantaged communities. We believe that ridesourcing policymakers should consider incorporating our proposed method into the travel demand modeling framework to inform fairer ridesourcing resource allocations and operations. Additionally, two fairness metrics can be used by city governments to evaluate and regulate ridesourcing operations. Moreover, the fairness measurements and unfairness correction method can be adopted to facilitate the effective operations of other travel modes such as public transit and shared micromobility. For example, an accurate and fair demand forecasting model will enable transit authorities to provide more personalized transit services to balance operation efficiency and effectiveness \citep{ermagun2020equity}. Also, a fairness-aware travel demand forecasting model will help micromobility (e.g., bikeshare and e-scooter) operators better rebalance the vehicles and ensure fair distribution of service availability throughout the day \citep{yan2021spatiotemporal}.


\subsection{Limitations and future research directions}

This study has some limitations that warrant follow-up investigations. For example, we only evaluated the proposed methodology using two fairness metrics (i.e., prediction accuracy gap and correlation coefficient) in this paper. Future works may consider using a wider range of fairness metrics to conduct a comprehensive evaluation. Moreover, by using correlation techniques, we assume the prediction accuracy is linearly correlated with the protected attributes. Future studies may consider exploring whether this association is nonlinear and developing corresponding methods. Another widely debated research topic is the connection between accuracy and fairness. Several previous studies have shown that the accuracy-fairness trade-off exists across datasets and applications \citep{berk2017convex, chouldechova2018frontiers} while others have shown that improvements in accuracy and fairness can co-occur \citep{yan2021fairness}. Hence, forthcoming investigations may shed further light on this relationship, such as identifying scenarios in which fairness and accuracy can both be enhanced or where the accuracy-fairness trade-off is prominent. Finally, this study only examined one travel mode (i.e., ridesourcing). A more comprehensive analysis that includes various travel modes (e.g., transit, car-sharing, and shared micromobility) and diverse contexts (e.g., different locations) should be conducted to test the generalizability and robustness of the unfairness correction method.

\section{Conclusion}
\label{conclusion}

This study examines the fairness issues in travel demand forecasting models and develops a new methodology to enhance their fairness while preserving the prediction accuracy. By leveraging two real-world ridesourcing-trip data from Chicago, IL and Austin, TX, the unfairness issues of seven state-of-the-art AI-based models on forecasting travel demand are evaluated. A novel and transparent in-processing method, which is based on an absolute correlation regularization term, is proposed to simultaneously address the unfairness arising from multiple protected attributes. We also compare the performance (including both fairness and accuracy) of our proposed unfairness correction method with three state-of-the-art unfairness correction method to show its effectiveness.

The results highlight that both statistical and machine learning models have pronounced fairness issues, i.e., the prediction accuracy for advantaged groups are notably higher than disadvantaged groups. Our proposed unfairness correction method can effectively enhance fairness for multiple protected attributes while preserving prediction accuracy. The comparative study reveals that our proposed method significantly outperforms other methods in both fairness and accuracy. Besides performance, our proposed method has remarkable flexibility---it is model-agnostic and can be adapted to different applications and different data modality. In summary, this study advances our understanding of fairness issues in travel demand forecasting and equips transportation researchers with a powerful tool to foster fairness within the transportation ecosystem.


\section*{Authorship Contribution Statement}
The authors confirm contributions to the paper as follows: \textbf{Zhang}: Conceptualization, Data Curation, Methodology, Software, Formal Analysis, and Draft Preparation. \textbf{Ke}: Conceptualization, Methodology, Formal Analysis, Draft Preparation. \textbf{Zhao}: Conceptualization, Methodology, Draft Preparation, Supervision and Grant Acquisition.


\section*{Acknowledgment}

This research was partially supported by the U.S. Department of Transportation through the Southeastern Transportation Research, Innovation, Development and Education (STRIDE) Region 4 University Transportation Center (Grant No. 69A3551747104) and through the Tier 1 University Transportation Center for Equitable Transit-Oriented Communities (CETOC) (Grant No. 69A3552348337). During the preparation of this work the authors used ChatGPT in order to check grammar errors and improving languages. After using this tool, the authors reviewed and edited the content as needed and takes full responsibility for the content of the publication.




\appendix
\label{Appendix}


\section{}
\label{appendix_a}

The following two figures show the spatial distribution of the protected attributes and the average ridesourcing demand per hour for two case studies.

\setcounter{figure}{0}
\renewcommand{\thetable}{A.\arabic{figure}}

\begin{figure}[h]
    \centering
    \includegraphics[width=\textwidth]{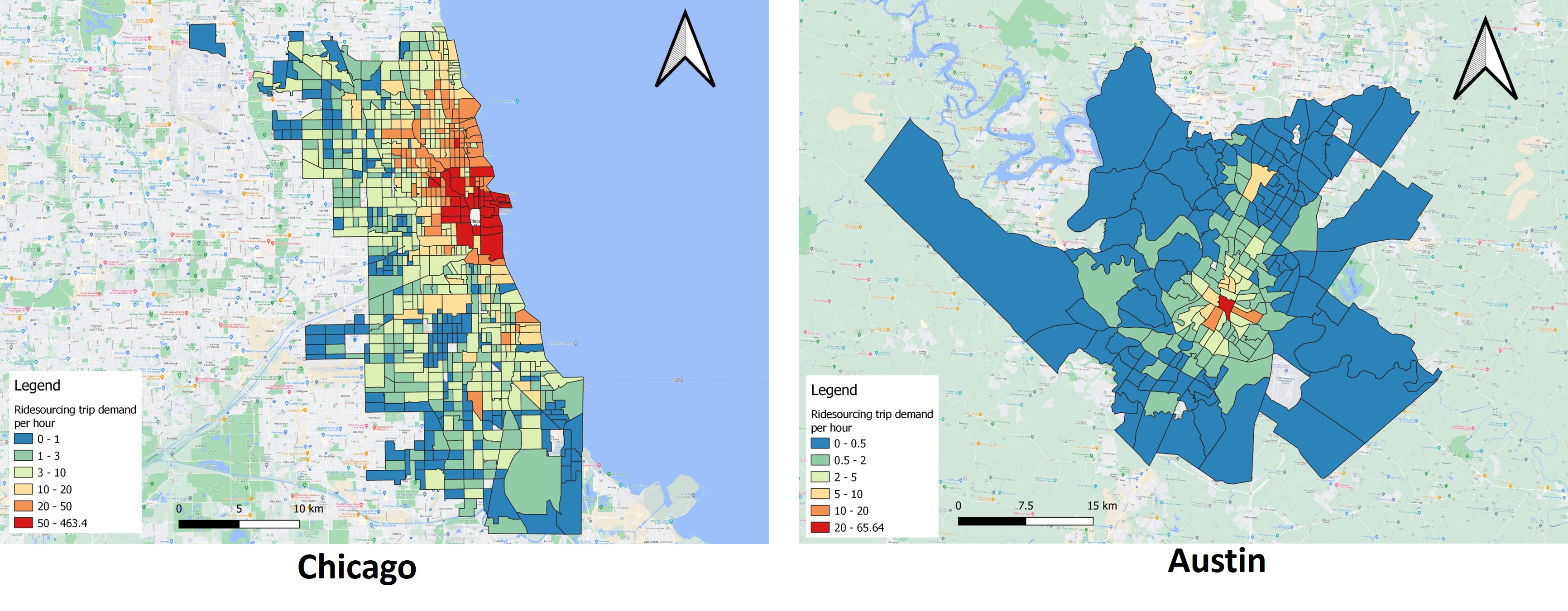}
    \caption{Spatial distribution of the average ridesourcing demand per hour}
    \label{fig:ridesourcing demand}
\end{figure}

\section{}
\label{appendix_b}

The correlation matrix of the selected four protected attributes for two case studies is shown in Table. \ref{tab:corr_matrix}. Results show that the protected attributes are evidently correlated with each other.

\setcounter{table}{0}
\renewcommand{\thetable}{B.\arabic{table}}

\begin{table}[]
\centering
\caption{Correlation matrix of protected attributes in two case studies}
\label{tab:corr_matrix}
\resizebox{0.8\textwidth}{!}{%
\begin{tabular}{@{}lllll|llll@{}}
\toprule
       & \multicolumn{4}{l|}{Chicago}      & \multicolumn{4}{l}{Austin}        \\ \midrule
       & Race   & Edu    & Age    & Income & Race   & Edu    & Age    & Income \\ \midrule
Race   & 1.000  & 0.620  & 0.504  & -0.748 & 1.000  & 0.605  & -0.178 & -0.401 \\
Edu    & 0.620  & 1.000  & 0.682  & -0.610 & 0.605  & 1.000  & -0.062 & -0.424 \\
Age    & 0.504  & 0.682  & 1.000  & -0.403 & -0.178 & -0.062 & 1.000  & 0.620  \\
Income & -0.748 & -0.610 & -0.403 & 1.000  & -0.401 & -0.424 & 0.620  & 1.000  \\ \bottomrule
\end{tabular}%
}
\end{table}


\section{}
\label{appendix_c}

The results of the sensitivity analysis for fairness weight, i.e., $\lambda$ are presented in Fig. \ref{fig:sensitivity_analysis_final}. We specifically investigated the effects of $\lambda$ on model's prediction accuracy by RMSE and fairness by both PAG and correlations. Note: In Austin case, setting fairness weight as 0.4 for GRU led to a substantial increase in RMSE and PAG. One possible reason could be that this combination of hyperparameters might explode the gradients and thus lead to this numerical instability.

\setcounter{figure}{0}
\renewcommand{\thetable}{C.\arabic{figure}}

\begin{figure*}[!ht]
    \centering
    \includegraphics[width = \textwidth]{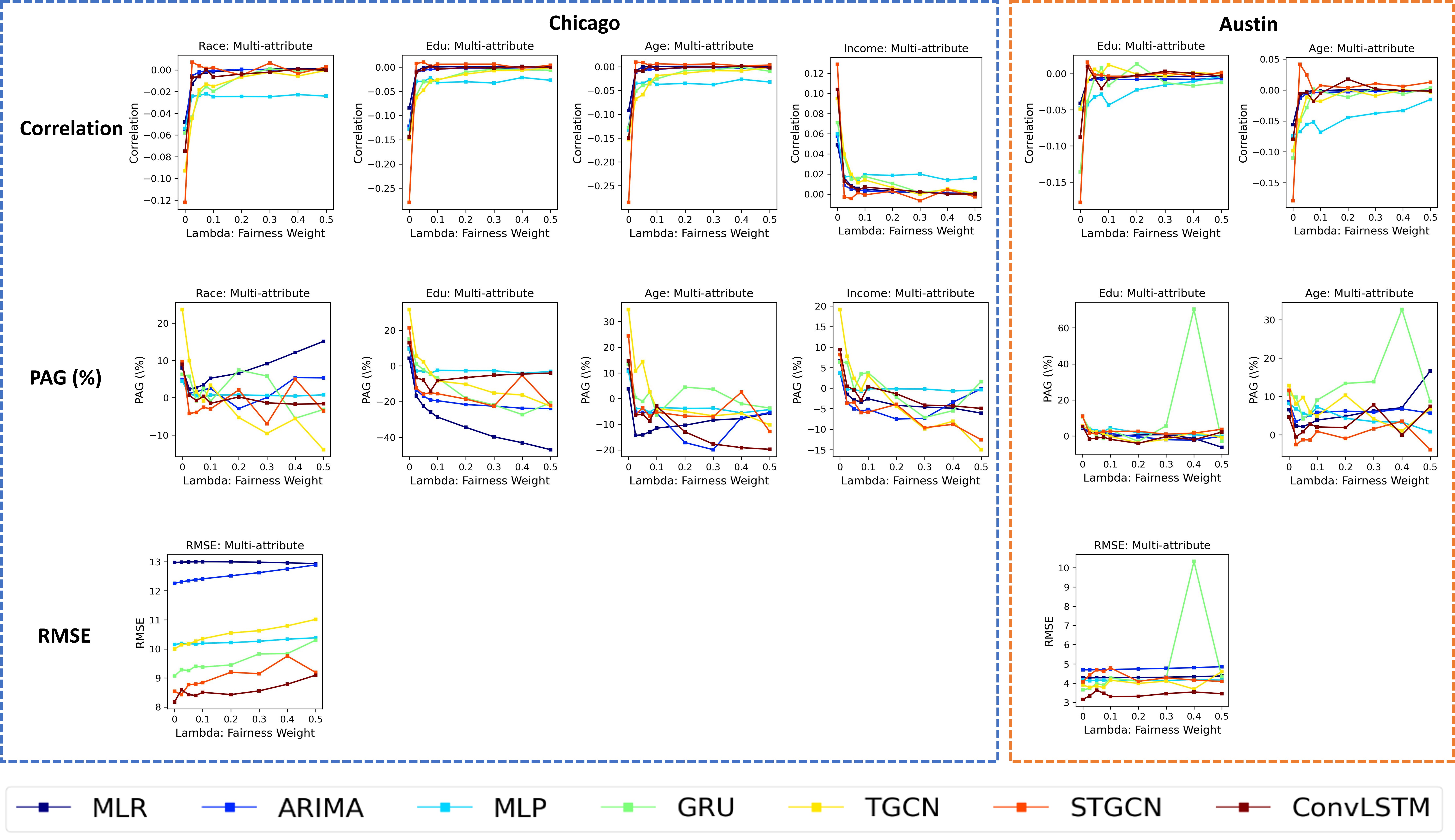}
    \caption{Sensitivity analysis of $\lambda$ across two case studies}
    \label{fig:sensitivity_analysis_final}
\end{figure*}

\FloatBarrier
\newpage






\bibliographystyle{elsarticle-harv}
\biboptions{semicolon,round,sort,authoryear}
\bibliography{sample.bib}







\end{document}